\documentclass[]{interact}

\usepackage{graphicx}
\usepackage{xcolor}
\usepackage{amsmath}
\usepackage{amssymb}
\usepackage{algorithm}
\usepackage{algorithmicx}
\usepackage[noend]{algpseudocode}
\usepackage{booktabs}
\usepackage[titletoc]{appendix}
\usepackage{titlesec}
\usepackage{listings}
\usepackage{enumitem}
\usepackage{multirow}
\usepackage{placeins}
\usepackage{rotating}
\usepackage{pifont}
\usepackage{array}
\usepackage{mdframed}
\usepackage{lipsum}
\usepackage[utf8]{inputenc}
\usepackage[many]{tcolorbox}
\usepackage{varwidth}
\usepackage{colortbl}
\usepackage{array}
\usepackage[colorlinks=true, allcolors=blue]{hyperref}
\usepackage{cleveref}
\usepackage{makecell}

\newcommand{\Sajad}[1]{\textcolor{black}{#1}}%orange
\newcommand{\SajadT}[1]{\textcolor{black}{#1}}%blue

\newcommand{\Bruce}[1]{\textcolor{black}{#1}}%orange

\usepackage[numbers,sort&compress]{natbib}
\bibpunct[, ]{[}{]}{,}{n}{,}{,}
\renewcommand\bibfont{\fontsize{10}{12}\selectfont}
\makeatletter\def\NAT@def@citea{\def\@citea{\NAT@separator}}
\makeatother

\usepackage[caption=false]{subfig}% Support for small, `sub' figures and tables
\title{
\fontsize{13.75}{13.75} \selectfont
\textcolor{black}{Systematic Comparison of Path Planning Algorithms using PathBench}}

\author{
\name{
Hao-Ya Hsueh\textsuperscript{a} Alexandru-Iosif Toma\textsuperscript{b}
Hussein Ali Jaafar\textsuperscript{a} 
Edward Stow\textsuperscript{b} \\
Riku Murai\textsuperscript{b} 
Paul H.J. Kelly\textsuperscript{b} and 
Sajad Saeedi\textsuperscript{a}}
\affil{\textsuperscript{a}Ryerson University, 350 Victoria St, Toronto, Canada;\\\textsuperscript{b}Imperial College London, Exhibition Rd, South Kensington, London, UK}
}
\begin{document}
\maketitle

%===============================================================================

%%%%%%%%%%%%%%%%%%%%%%%%%%%%%%%%%%%%%%%%%%%%%%%%%%%%%%%%%%%%%
%%%%%%% Abstract
%%%%%%%%%%%%%%%%%%%%%%%%%%%%%%%%%%%%%%%%%%%%%%%%%%%%%%%%%%%%%
\begin{abstract}

\Bruce{Path planning is an essential component of mobile robotics. Classical path planning algorithms, such as wavefront and rapidly-exploring random tree (RRT) are used heavily in autonomous robots. With the recent advances in machine learning, development of learning-based path planning algorithms has been experiencing a rapid growth. An unified path planning interface that facilitates the development and benchmarking of existing and new algorithms is needed. This paper presents PathBench, a platform for developing, visualizing, training, testing, and benchmarking of existing and future, classical and learning-based path planning algorithms in 2D and 3D grid world environments. Many existing path planning algorithms are supported; e.g. A*, Dijkstra, waypoint planning networks, value iteration networks, gated path planning networks; and integrating new algorithms is easy and clearly specified. The benchmarking ability of PathBench is explored in this paper by comparing algorithms across five different hardware systems and three different map types, including built-in PathBench maps, video game maps, and maps from real world databases. Metrics, such as path length, success rate, and computational time, were used to evaluate algorithms. Algorithmic analysis was also performed on a real world robot to demonstrate PathBench's support for Robot Operating System (ROS). PathBench is open source\footnote{\href{https://sites.google.com/view/PathBench}{https://sites.google.com/view/PathBench}}. }    
 
\begin{comment}
Path planning is a key component in mobile robotics. A wide range of path planning algorithms exist, but few attempts have been made to benchmark the algorithms holistically or unify their interface. Moreover, with the recent advances in deep neural networks, there is an urgent need to facilitate the development and benchmarking of such learning-based planning algorithms. This paper presents PathBench, a platform for developing, visualizing, training, testing, and benchmarking of existing and future, classical and learned 2D and 3D path planning algorithms, while offering support for Robot Operating System (ROS). Many existing path planning algorithms are supported; e.g. A*, wavefront, rapidly-exploring random tree, value iteration networks, gated path planning networks; and integrating new algorithms is easy and clearly specified. We demonstrate the benchmarking capability of PathBench by comparing implemented classical and learned algorithms for metrics, such as path length, success rate, computational time and path deviation \textcolor{black}{on various hardware systems}. These evaluations are done on built-in PathBench maps and external path planning environments from video games and real world databases.
%\footnote{\href{https://github.com/perfectly-balanced/PathBench}{https://github.com/perfectly-balanced/PathBench}}.
\end{comment}

\end{abstract}
 

%===============================================================================

%%%%%%%%%%%%%%%%%%%%%%%%%%%%%%%%%%%%%%%%%%%%%%%%%%%%%%%%%%%%%
%%%%%%% Introduction
%%%%%%%%%%%%%%%%%%%%%%%%%%%%%%%%%%%%%%%%%%%%%%%%%%%%%%%%%%%%%
\section{Introduction}
\begin{figure}
\centering
{%
\resizebox*{7.5cm}{!}{\includegraphics{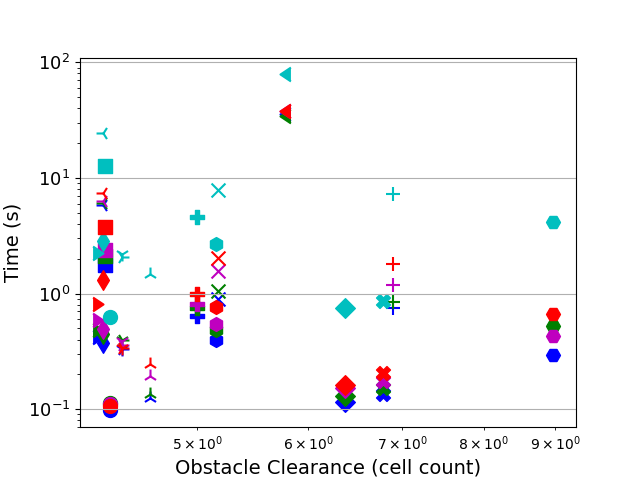}}}\hspace{0.2pt}
{%
\resizebox*{6.5cm}{!}{\includegraphics{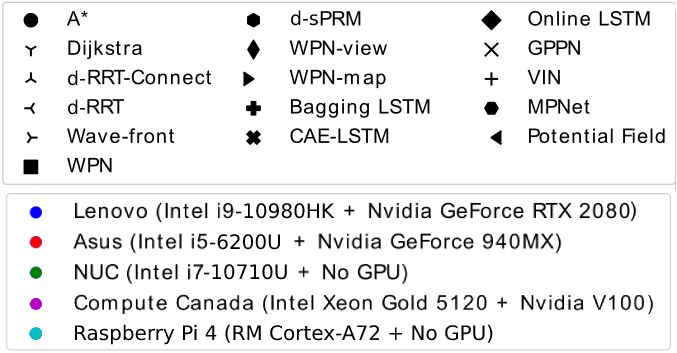}}}
\vspace{1mm}
\caption{\SajadT{Scatter plot from comparative studies using various hardware systems that demonstrates obstacle clearance (higher is better) and time performance (lower is better) of algorithms. It is observed that obstacle clearance is consistent across different hardware systems and computation time of algorithms is faster in systems with more powerful CPU and GPU. \Sajad{To generate this figure, each algorithm, represented by a shape, was run on 3000 maps of size 64$\times$64, and the results were averaged.} Additional analysis with plots generated by PathBench can be found in Sec.~\ref{sec:hard}.}}
\label{fig: scatter1}
\end{figure}

%Path planning is an optimization problem with one or multiple objectives that aims 
\Sajad{In robotics, path planning remains as an open problem, as a multi-objective optimization problem, }
\Bruce{to generate a feasible and continuous path that connects a system from its start to goal configuration.} 
%
%Autonomous robots rely on various path planning algorithms to meet specific performance metrics \cite{gonzalez2016review}.
\Sajad{Various algorithms exist, yet the research on developing newer ones is still ongoing \cite{gonzalez2016review,lavalle2006planning,choset2005principles}. In particular, recently there is rapid growths in methods that rely on machine learning and deep neural networks, allowing to solve complex planning problems.}
%
%In robotics and motion planning, benchmarking and comparison between algorithms is key to the experimental evaluation of newly proposed algorithms. Papers often report performance scores such as length or curvature of the path, but also increasingly complex metrics, such as execution time and memory consumption. However, as the diversity of the algorithms is expanding, especially with the advances in the areas of machine learning (ML), it becomes a challenging issue to benchmark algorithms efficiently. 
\Sajad{This indicates the urgent need the community have to develop a unified benchmarking framework to compare the existing and new algorithms against metrics such as length and smoothness of paths, run time and memory consumption, and the success rate of the algorithms.}
%
%To address this issue, we present a unified framework that supports benchmarking and development of classical and learned planning algorithms (See Fig.~\ref{fig: sim}). 
\Sajad{In this work, we aim to solve this problem by presenting a modular simulation \& benchmarking platform and conducting a comparative study, demonstrating how various algorithms on different hardware compare considering the metrics of interest. (See Fig.~\ref{fig: scatter1} and Fig.~\ref{fig: sim_platform})}
\par
\SajadT{
We compare various classical and machine-learning based path planning algorithms across a range of hardware, typically used in robotic applications, that produced Fig.~\ref{fig: scatter1}. Extensive results are presented in Section~\ref{sec:result}. The results presented here are easily reproducible and extendable to other existing and future algorithms, using the unified platform, coined PathBench. These results are important in applications that seek Pareto optimality or require trades-offs between several performance metrics, such as execution time, path length, and trajectory smoothness.} \SajadT{For instance, from Fig. \ref{fig: scatter1}, it is evident that A* is the fastest algorithm; however, it does not optimize for obstacle clearance. In contrast, VIN and MPNet, which are neural network-based planners, produce paths that are not nearly as close to obstacles.}

%%%%%%%%%%%%%%%%%%%%%%%%%%%%%%%%%%%%%%%%%%%%%%%%%%%%%%%%%%%%%
%%%%%%% Figure Pareto front 2 Systems
%%%%%%%%%%%%%%%%%%%%%%%%%%%%%%%%%%%%%%%%%%%%%%%%%%%%%%%%%%%%%
%\begin{figure*}[ht!]
%    \centering
%    \includegraphics[width=1\textwidth]{images/pareto_small.png}
%    \caption{\textcolor{black}{Comparative results of several classical and learned path planning algorithms across various hardware.}}
%    \label{fig:my_label}
%        \vspace{-5 mm}
%\end{figure*}

%
%In this paper we introduce 

\Sajad{This comparative study presented utilizes PathBench}, a motion planning platform that can be used to develop, assess, compare, and visualize the performance and behaviour of path planning algorithms. \Sajad{PathBench currently supports only grid map environments; however, it will be expanded to include continuous space and topological maps and algorithms in the future.} The key contributions of this work include: 
\begin{itemize}
\item Creation of an unified path planning development and benchmarking platform, that supports both 2D and 3D classical and learned algorithms. Existing machine learning based algorithms, such as value iteration networks (VIN) \cite{tamar2016value}, gated path planning networks (GPPN) \cite{lee2018gated}, 
motion planning networks (MPNet) \cite{qureshi2019motion}, as well as Online LSTM~\cite{nicola2018lstm}, and CAE-LSTM~\cite{inoue2019robot} methods, are incorporated into PathBench. PathBench has a structured environment to facilitate easy development and integration of new classical and ML-based algorithms. The platform provides interfaces for algorithm visualization, rapid development, training, training data generation and benchmarking analysis. Waypoint Planning Networks~(WPN)~\cite{wpnconf} is an algorithm developed within PathBench to showcase its feasibility.
\item PathBench's benchmarking features allow evaluation against the suites of added path planning algorithms, both classical algorithms and machine-learned models, with standardized metrics and environments.
\item PathBench provides a ROS (Robot Operating System) real-time extension for interacting with a real-world robot. Examples have been provided on the GitHub of the project.
\item \SajadT{Using PathBench, we provide a comparative analysis of a large suite of publicly available classical and learning-based path planning algorithms across different map types, datasets and hardware systems for point mass in 2D and 3D grid environments.\textcolor{black}{\footnote{This paper is partly based on our earlier conference paper~\cite{PB_CRV_2021}, extended with \Sajad{across hardware/algorithm} benchmarking results.}}} 
\end{itemize}
\section{Related Work}
In this section, classical and learned planning algorithms, and existing \Bruce{simulation \&} benchmarking frameworks are reviewed briefly.

%\subsection{Classical and Learned Planning Algorithms}
\subsection{\Sajad{Path Planning Algorithms}}
\Sajad{Based on map representation and algorithmic processing, there are various groups of path planning algorithms. For instance, graph search algorithms, such as Dijkstra \cite{choset2005principles},  A* \cite{duchovn2014path}, and wavefront \cite{luo2014effective}, commonly applied to grid and lattice maps, generate optimal results but are slow at applications with high-dimensional spaces such as robotic manipulators with many degrees of freedom. Sampling-based algorithms, such as rapidly-exploring random tree (RRT) \cite{lavalle1998rapidly} and probabilistic roadmap (PRM) \cite{kavraki1994probabilistic}, deal with the curse of dimensionality by sampling the configuration space or the state space to generate a path. However, these algorithms produce sub-optimal results. % %%%%
While graph search and sampling-based algorithms process the whole map, sensor-based planning algorithms, such as Bug1 and Bug2~\cite{choset2005principles, rajko2001pursuit} plan only for a local view~\cite{sensor1997, Paull_TMech_2013}. This way the sensory data is taken into account. % %%%%
Numerical optimization  algorithms, such as ~\cite{pot1992}, can also produce optimal results; however, they may become trapped in local minima. These algorithms optimize a cost function composed of kinematics constraints \cite{ziegler2014making} or obstacle clearance and trajectory smoothness such as STOMP~\cite{STOMP_2011}, CHOMP~\cite{CHOMP_2013}, and TrajOpt~\cite{TrajOpt_2014}. Some algorithms, such as \cite{dolgov2010path}, uses a combination of grid search and numerical optimization.
Another important class of algorithms is the data-driven learning-based algorithms~\cite{Schwarting2018PlanningAD}. This algorithm can solve complex problems utilizing recent advances in high performance computing algorithms and hardware~\cite{inoue2019robot,Chen2016Humanoids,gupta2017cognitive, nicola2018lstm, qureshi2018deeply, chamzas2019using, qureshi2018motion, CoMPNet, qureshi2019motion, bency2019neural, TDPPNet, mohammadi2018path, choi2020pathgan, tamar2016value, lee2018gated, srinivas2018universal, Levine2013, Abbeel2010}.}

\subsection{Simulation and Benchmarking}% Platforms}
{Benchmarking} of algorithms is the scientific approach of evaluation in the robotics community. \Bruce{In order to generate path planning benchmarks, reproducible experiments are required, which are easier in simulated environments. %Benchmarking of path planning algorithms are mostly performed by evaluating generated robot paths from one location to another. 
However, a common set of metrics, maps, and simulation environments has not been adopted for the evaluation of classical and learning-based path planning algorithms.}

\par
\SajadT{Currently, there is a variety of libraries relevant to the simulation and benchmarking of path planning algorithms, such as \textit{OpenRAVE} \cite{diankov2008openrave}, \textit{OMPL} \cite{sucan2012the_open_motion_planning_library},  \textit{MoveIt} \cite{moveit}, \textit{SBPL} \cite{plaku2007oops}, and \textit{OOPS\textsubscript{MP}} \cite{plaku2007oops}.} \Bruce{Most of these platforms have been designed for simulation and development of new planning methods. The ability to benchmark algorithms is also incorporated into OOPS\textsubscript{MP}, MoveIt, and OMPL to facilitate algorithmic analysis~\cite{omplbench,plaku2007oops}, but they do not support the evaluation and development of learning-based algorithms.} \textcolor{black}{Table~\ref{tab: plat_comparison} compares the key features of these frameworks with PathBench. }

\begin{table}[t]
    \footnotesize
    \center
    \caption{Capabilities comparison \Bruce{for existing frameworks that focus on path planning algorithms.} PathBench supports \Bruce{development, visualization, and} benchmarking of classical and learned planning algorithms. %H - High, M - Moderate, R - Reduced. ECI is short for Environment Complexity and Interaction
    }
    \vspace{1mm}
    \begin{tabular}{|c| c c | c c c |}% c c c |}
         \hline
         {\scriptsize \rotatebox[origin=c]{0}{Platform}} & {\scriptsize \rotatebox[origin=c]{60}{Visualization}} & {\scriptsize \rotatebox[origin=c]{60}{Benchmarking}} & {\scriptsize
         %\rotatebox[origin=c]{60}{\textcolor{red}{Custom Algorithm}}} & {\scriptsize
         \rotatebox[origin=c]{60}{Sample-Based}} &{\scriptsize \rotatebox[origin=c]{60}{Graph-Based}} &  {\scriptsize \rotatebox[origin=c]{60}{ML-Based}} \\%&  {\scriptsize \rotatebox[origin=c]{90}{Efficiency}} & {\scriptsize \rotatebox[origin=c]{90}{Variety}} & {\scriptsize \rotatebox[origin=c]{90}{ECI}} \\
         %\hline
         %{ROS} & \textcolor{green}{\checkmark} & \Bruce{$\times$} & %\textcolor{green}{\checkmark}& \textcolor{green}{\checkmark} & %\textcolor{green}{\checkmark} & \Bruce{$\times$} \\% & {R} & {H} & {H} \\
         \hline
         {OpenRAVE} & \textcolor{green}{\checkmark} & \textcolor{red}{$\times$} &  \textcolor{green}{\checkmark} & \textcolor{red}{$\times$} & \textcolor{red}{$\times$}  \\% {M} & {H} & {H} \\
         \hline
         {OMPL} & \textcolor{green}{\checkmark} & \textcolor{green}{\checkmark} & \textcolor{green}{\checkmark} & \textcolor{red}{$\times$} & \textcolor{red}{$\times$} \\% & {H} & {R} & {R} \\
         \hline
         {MoveIt} & \textcolor{green}{\checkmark} & \textcolor{green}{\checkmark} &  \textcolor{green}{\checkmark} & \textcolor{red}{$\times$} & \textcolor{red}{$\times$} \\% & {M} & {H} & {H} \\
         \hline
         {SBPL} & \textcolor{green}{\checkmark} & \textcolor{red}{$\times$} &  \textcolor{red}{$\times$} & \textcolor{green}{\checkmark} & \textcolor{red}{$\times$}  \\% {R} & {R} & {M} \\
         \hline
         {OOPS\textsubscript{MP}} & \textcolor{green}{\checkmark} & \textcolor{green}{\checkmark} & \textcolor{green}{\checkmark} & \textcolor{red}{$\times$} & \textcolor{red}{$\times$} \\% & {M} & {M} & {M} \\
         \hline
         \textbf{PathBench} & \textcolor{green}{\checkmark} & \textcolor{green}{\checkmark} &  \textcolor{green}{\checkmark} & \textcolor{green}{\checkmark} & \textcolor{green}{\checkmark} \\% & {H} & {M} & {R} \\
         \hline
    \end{tabular}
    \label{tab: plat_comparison}
        \vspace{-5 mm}
\end{table}

%The {analyzer} is quite involved and extensible, but the  
%{Simulator} has limited capabilities in both rendering and environment interaction compared to the other libraries. 

%The main advantages of PathBench over existing standardized libraries is its native support of ML methods, as well as its simple, lightweight, and extensible design, allowing fast prototyping and benchmarking for an ideal research environment. Moreover, we provide a clean API interface for the algorithms which makes them portable to the standardized libraries. Also, we provide a \textit{ROS} real-time extension which converts the internal map move actions into network messages (velocity control commands) using the {ROS} publisher-subscriber APIs. See Table \ref{tab: plat_comparison} for platform comparison. and not thethat the platform was built to be used in an ideal research environment. Since, we use ML methods, the focus was set on the path generation and not the interaction between the environment and the agent

\SajadT{It is noticed that most existing benchmarking and simulation platforms do not support machine learning algorithms. However, newer projects such as iGibson~\cite{shenigibson}, Habitat~\cite{habitat19iccv}, and PyRobot~\cite{pyrobot} have started focusing on providing environments for artificial intelligence research in robotics.} 
\begin{comment} %Sajad removed this for the 6000 word limit.
iGibson provides a large and realistic simulation environment for development of solutions for interactive tasks in robotics. Habitat is a platform for research in embodied artificial intelligence that encompasses photorealistic 3D simulation. PyRobot is a lightweight robotics framework built on top of ROS that is designed for research and benchmarking in high-level AI applications. It is hardware independent and allows control of various robots. BARN provides a suite of simulation environment that supports learning-based navigation to test low-level motion skills of mobile robot in cluttered environments.
\end{comment}

\textcolor{black}{PathBench also improves on previously developed platforms, \Bruce{but focuses presently on path planning development and benchmarking in grid space environments.}} The main advantages of PathBench over existing standardized libraries are its native support of machine learning path planning algorithms, as well as its simple and lightweight design, allowing fast prototyping for a research environment. \textcolor{black}{A clean API for the algorithms also allows PathBench to be portable to standardized libraries and integration into iGibson and Habitat is possible. In addition, PathBench’s support for animated simulations and its simple interface allow it to be a suitable educational tool.} It provides a standardized set of maps and metrics, so that benchmarking of new and existing algorithms can be performed quickly. We also provide a {ROS} real-time extension which converts the internal map move actions into network messages (velocity control commands) using the 
{ROS} APIs. 

Further to these standard libraries, other works have sought to address the issue of benchmarking for motion planning. Althoff {\it et al.} provide a collection of benchmarks for motion planning of cars on roads, allowing reproducible results on problems~\cite{althoff2017commonroad}. \Bruce{By using OMPL and MoveIt, a benchmark of sampling-based algorithms used for grasping tasks of three manipulators has been formulated~\cite{grasping}. Benchmark for Autonomous Robot Navigation (BARN) is composed of ground navigation scenarios with emphasis on online path planning approaches, such as Dynamic Window Approach (DWA) and Elastic Bands
(E-Band) ~\cite{perille2020benchmarking,dwa,eband}. PathBench currently focuses instead on support of offline path planning methods in grid environments. On the other hand,} Sturtevan provides a standard test set of maps, and suggests standardized metrics for grid based planning for gaming environments~\cite{sturtevant2012benchmarks}. These maps have been ported into PathBench and are discussed further in Sec.~\ref{sec:maps}.  

\Bruce{Although comparative studies between classical path planning algorithms have been done for UAV and mobile robot navigation \cite{uavcomp,robcomp}, studies that compare both classical and learning-based path planning algorithms across different hardware systems \Sajad{are rare}. Evaluations of algorithms on multiple hardware systems is especially essential in the field of robotics, due to potential size, weight, and power consumption constraints of embedded hardware devices for mobile robotics applications. \Sajad{For other robotics research problem, such as visual odometry and SLAM, there exist such studies. For instance,} Delmerico and Scaramuzza have performed a benchmark comparison of monocular visual-inertial odometry algorithms across embedded hardware systems that are commonly adopted for flying robots \cite{visualodom}. \Sajad{SLAMBench series of papers conduct similar studies for a wide range of algorithms across various hardware \cite{SLAMBench, SLAMBench2, SLAMBench3}.} Performance of path planning algorithms across commonly utilized hardware systems in robotics applications should also be examined, so that system-specific algorithmic benchmarks can be developed. PathBench facilitates cross-system evaluations and allows users to improve and select the appropriate algorithm for a desired task. In this paper, we perform a comparative study using PathBench and its feature for classic and learned planning algorithms. Fig. \ref{fig:4scatter} showcases scatter plots that demonstrate trade-offs between algorithms across five different hardware configurations.}

%\subsection{Contributions}
%\textcolor{black}{The following contributions are made by this paper:}
%\par
%\begin{itemize}
%    \item \textcolor{black}{Creation of an unified path planning development and benchmarking platform that supports both classical and machine learning-based algorithms. The platform provides interfaces for algorithm visualization, rapid development, training, training data generation and benchmarking analysis. Waypoint Planning Networks~(WPN)~\cite{wpnconf} is an algorithm developed within PathBench to showcase its feasibility.} 
%    \item \textcolor{black}{Comparative analysis of a large suite of publicly available classical and learning-based path planning algorithms across different map types, datasets and hardware systems for point mass in 2D and 3D environments.}
%\end{itemize}

%%%%%%%%%%%%%%%%%%%%%%%%%%%%%%%%%%%%%%%%%%%%%%%%%%%%%%%%%%%%
%%%%%%% PathBench Platform
%%%%%%%%%%%%%%%%%%%%%%%%%%%%%%%%%%%%%%%%%%%%%%%%%%%%%%%%%%%%%
\section{PathBench Platform}

\begin{figure}
    \centering
    %\hspace{-10 mm}
    \includegraphics[scale=0.4]{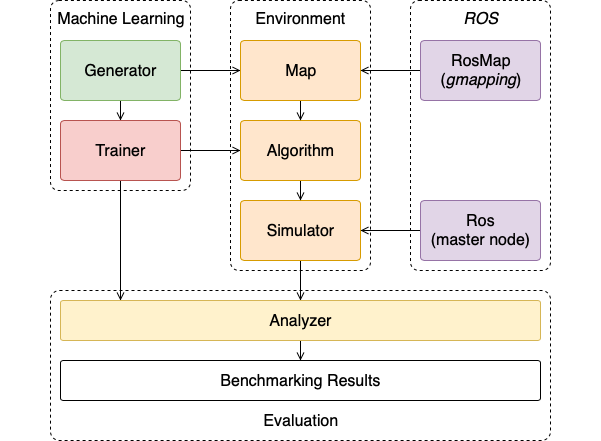}
     \vspace{1mm}
    \caption{PathBench structure overview. Arrows represent information flow/usage ($A \xleftarrow{gets/uses} B$). The machine learning section is responsible for training dataset generation and model training. The Environment section controls the interaction between the agent and the map, and supplies graphical visualization. The \textit{ROS} section provides support for real-time interaction with a real physical robot. The Evaluation section provides benchmarking methods for algorithm assessment.} %For a detailed architecture, please refer to the website of PathBench.}
    \label{fig: sim_platform}
        \vspace{-5 mm}
\end{figure}

An overview of the architecture of PathBench is shown in Fig. \ref{fig: sim_platform}. PathBench is composed of four main components: \emph{Simulator}, \emph{Generator}, \emph{Trainer}, and \emph{Analyzer} \Bruce{where infrastructures are created to link the four main components with other parts of the framework to} provide general service libraries and utilities.
%
%\textbf{Simulator.} This section 
The simulator is responsible for environment interactions and algorithm visualization. It provides custom collision detection systems and a graphics framework for rendering the internal state of the algorithms.
%
%\textbf{Generator.} This section 
The generator is responsible for generating and labelling the training data used to train the ML models.
%
%\textbf{Trainer.} This section 
The trainer is a class wrapper over the third party machine learning libraries. It provides a generic training pipeline based on the holdout method and standardized access to the training data.
%
%\textbf{Analyzer.} The final section 
Finally, the analyzer manages the statistical measures used in the practical assessment of the algorithms. Custom metrics can be defined, as well as graphical displays for visual comparisons. 
PathBench has been written in Python, and uses PyTorch \cite{paszke2017automatic} for ML. %and \textit{pygame} \cite{pygame} for rendering. 
%The other option for ML was TensorFlow~\cite{tensorflow2015_whitepaper}, but {PyTorch} was used since it is more flexible in debugging RNNs with variable size inputs~\cite{tensorflow_vs_pytorch}.

%\textbf{Notation.} 
%Throughout the paper, we will use \textit{italic font} for libraries, \textbf{bold font} for classes and \texttt{typewriter font} for code snippets/file names/functions/variables. For types, we use the \textit{python} type hinting system (e.g. a list of integers is defined as \texttt{List[int]}).

%%%%%%%%%%%%%%%%%%%%%%%%%%%%%%%%%%%%%%%%%%%%%%%%%%%%%%%%%%%%%
%%%%%%% Figure
%%%%%%%%%%%%%%%%%%%%%%%%%%%%%%%%%%%%%%%%%%%%%%%%%%%%%%%%%%%%%

\begin{figure}
  \centering
  \subfloat[A*]{%
   \resizebox*{5cm}{!}{\includegraphics{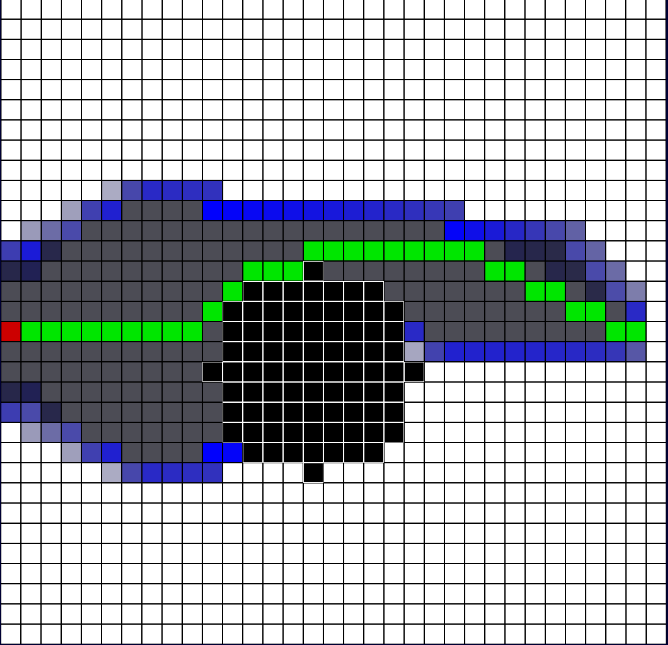}}}\hspace{5pt}
  \subfloat[Discrete RRT-Connect]{%
   \resizebox*{5cm}{!}{\includegraphics{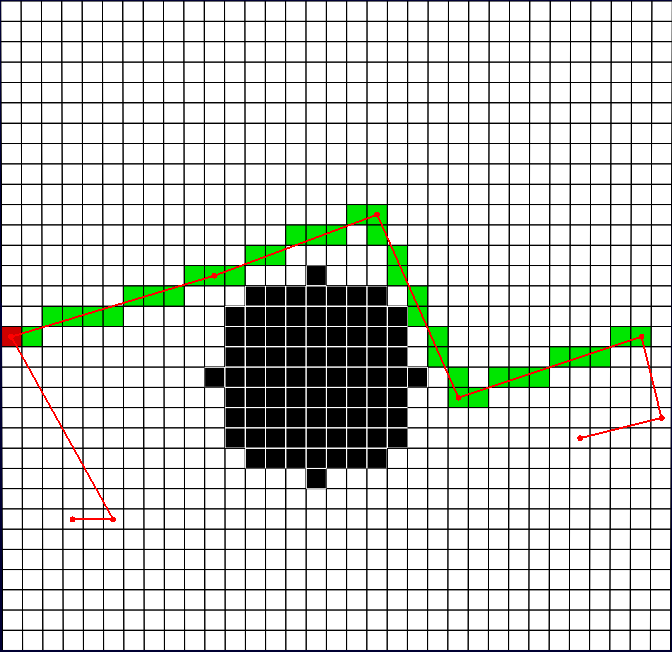}}}\hspace{5pt}
  \subfloat[GPPN]{%
   \resizebox*{5cm}{!}{\includegraphics{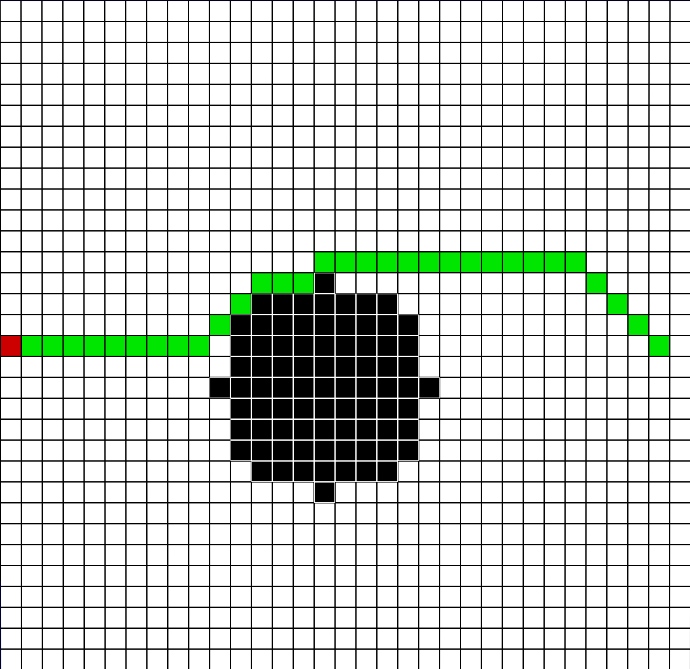}}}\hspace{5pt}
  \subfloat[WPN]{%
   \resizebox*{5cm}{!}{\includegraphics{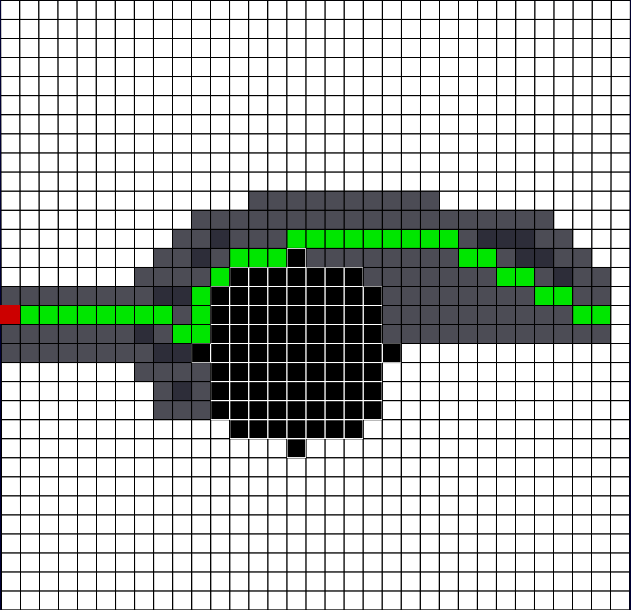}}}\hspace{5pt}
  \subfloat[ A*]{%
   \resizebox*{5cm}{!}{\includegraphics{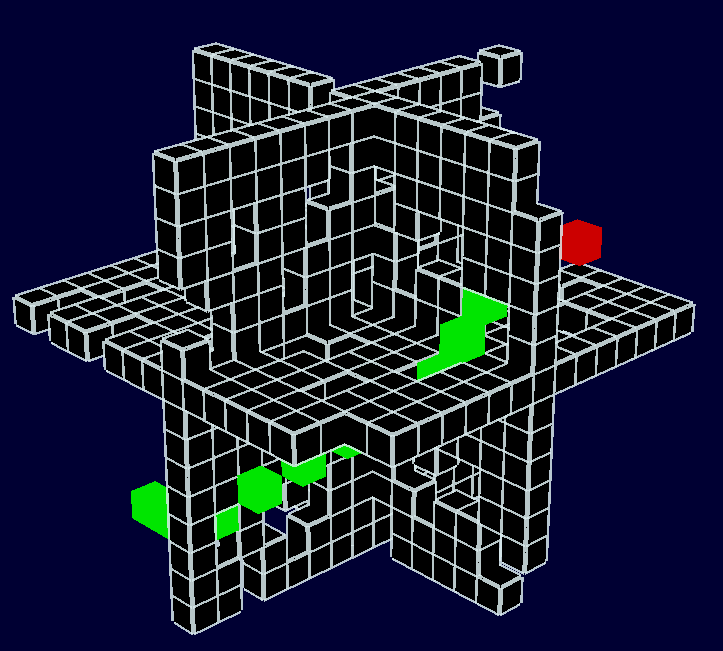}}}\hspace{5pt}
   \subfloat[Discrete RRT-Connect]{%
   \resizebox*{5cm}{!}{\includegraphics{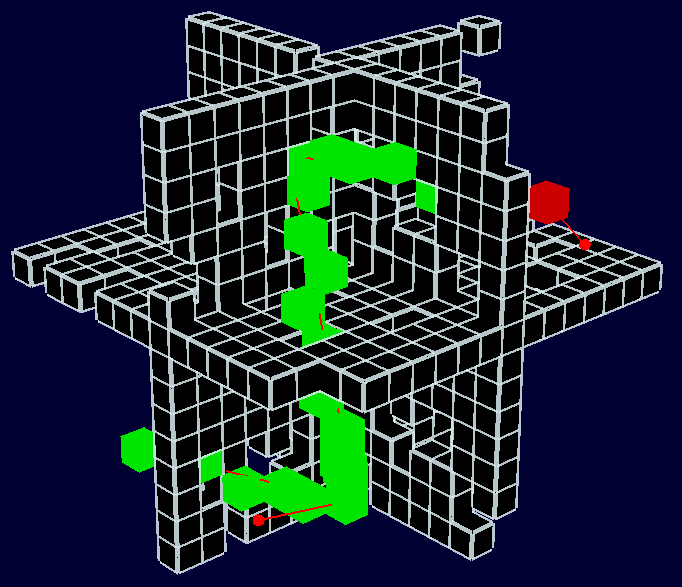}}}\hspace{5pt}
  \vspace{1 mm}
  \caption{The results of different classical and learned planners in PathBench. The red entity is the agent, light green cells show the path, with the green entity at the end of the path denoting the goal. The black entities are obstacles and everything else is custom display information \Bruce{(i.e. Dark gray represents the search space in (a) and (d), red represents edges of the sampling algorithm in (b), and blue shows the frontiers of the A* search space.) Discretized version of sampling-based algorithms~\cite{Morgan_IROS_2004} were used to ensure consistent performance in grid space.}}
  \label{fig: sim}
      \vspace{- 3mm}
\end{figure}

\subsection{Simulator}

\begin{figure}
    \centering
    %\hspace{-10 mm}
    \includegraphics[scale=0.2]{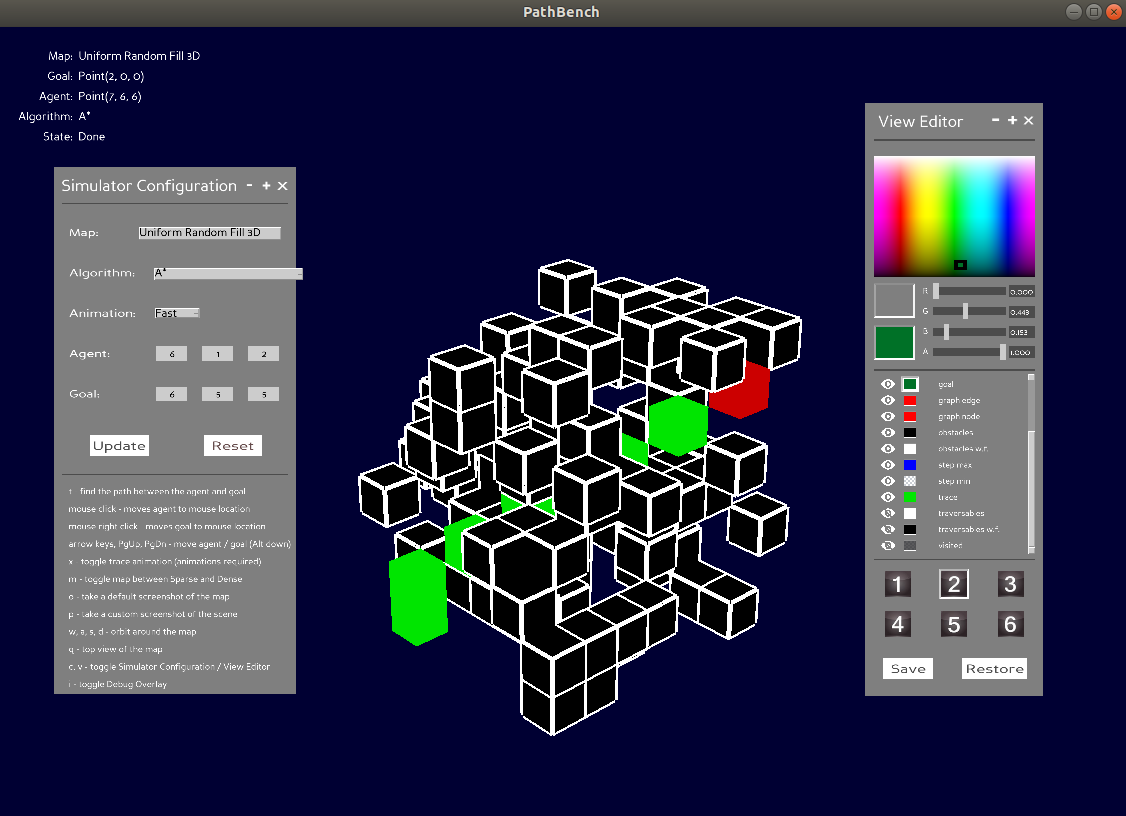}
     \vspace{1mm}
    \caption{ The GUI of the simulator is where configuration of the path planners, environments and customization of the visualization are made interactively. Simulator configuration window is used to set up path planning sessions\Bruce{, including the selection of algorithms, maps, and the start and goal points.} View editor allows for adjustment of \Bruce{the simulation's visualizations.}}
    \label{fig: simulator3d}
        \vspace{-5 mm}
\end{figure}

The {simulator} is both a visualizer and an engine for developing algorithms (Fig. \ref{fig: sim}). It supports animations and custom map display components which render the {algorithm}'s internal data. Simulator has a {map} that contains different entities such as the {agent}, {goal} and {obstacle}s, and provides a clean interface that defines the movement and interaction between them. Therefore, a {map} can be extended to support various environments; however, each map has to implement its own physics engine or use a third party one (e.g. the 
\textit{pymunk} physics engine 
%\cite{pymunk} 
or \textit{OpenAI Gym} 
%\cite{OpenAI_Gym}
). The current implementation supports three types of 2D/3D maps: {DenseMap}, {SparseMap} and {RosMap}, corresponding to static grid map, point cloud map, and grid map with live updates, respectively. Additionally, the {simulator} provides animations that are achieved through key frames and synchronization primitives. The graphical framework used for the visualization of planners and GUI is Panda3D \cite{panda}. Simulator configurations and visualization customizations can be directly controlled within the Panda3D GUI, see Fig.~\ref{fig: simulator3d}. 

\subsection{Generator} \label{sec: generator}

%The {generator} can execute five actions: (1) conversion from image to map, (2) map generation, (3) map labelling, (4) map augmentation, and (5) map modification.

The {generator} executes the following four actions, 
%: (1) map generation, (2) map labelling, (3) map augmentation, and (3) map modification, 
each explained briefly below.

%(1) \textbf{Conversion.} An image can be converted into an internal \textbf{Map} and saved into the maps directory from \textbf{Resources}. The image can be a software drawn image or Simultaneous Localization and Mapping (SLAM) image \cite{dissanayake2001solution} output from a real robot, as long as it follows the conventions imposed by the image converter: an agent represented by a true red circle has to be present, a goal represented by a true green circle has to be present and the obstacles need to be in the gray-black colour range. % (See Fig. \ref{fig: converted map}).

%\begin{figure}[h]
%  \centering
%  \begin{subfigure}[b]{0.48\linewidth}
%    \includegraphics[width=\linewidth]{images/map10.png}
%     \caption{Original SLAM Image}
%  \end{subfigure}
%  %\hspace{1.5cm}
%  \begin{subfigure}[b]{0.48\linewidth}
%    \includegraphics[width=\linewidth]{images/screenshot_108.png}
%     \caption{Converted \textbf{Map}}
%  \end{subfigure}
%  \caption{Example of the conversion process from the \textbf{Generator}}
%  \label{fig: converted map}
%\end{figure}

%\todo{Sajad: do you ever refer to Fig 4.2-h? it seeme like a  repetition of 4.3-b.}
% The repetion is used to hilight that the simulator has a grid display as well, even if it is the same algorithm

%\todo{Sajad: It is very common to use 'Figure X' instead of 'figure X' and 'Algorithm Y' instead of 'Algorithm Y' and 'Table Z' instead of 'table Z' when you want to refer a figure, alg, or table in the text.}

{\it 1) Generation.} The generation procedure accepts as input, different hyper-parameters such as the type of generated maps, number of generated maps, number of dimensions, obstacle fill rate range, number of obstacle range, minimum room size range and maximum room size range. %, which define the structure of the maps. 
%When a range is given as input, the generator picks a random number between the range and feeds it into the associated map type generator. 
Currently, the generator can produce four types of maps: uniform random fill map, 
%(See Algorithm \ref{alg: Uniform random fill generator}), 
block map, house map and point cloud map,
%(See Algorithm \ref{alg: Block map generator}) and 
%(See Algorithm \ref{alg: House generator}) 
%(See Fig. \ref{fig: generated maps}) 
(See Fig. \ref{fig:3maps}).
\begin{comment} % Sajad removed for 6000 word limit

and it can be extended to support different synthetic maps such as mazes and cave generation using cellular automata. All generated maps are placed into an {Atlas} directory in both .pickle and JSON formats. 
\end{comment}

%which is a custom directory that saves files using the index number. %It keeps track of the next available index, and thus, when a new file is saved, it is "appended" to it. \textbf{Atlas} directories are used for easier indexing operations such as index loading and index saving. %(the file system service is described in Appendix \ref{sec: infra}).

{\it 2) Labelling.} The labelling procedure takes a map 
\begin{comment} % Sajad 6000
from {Atlas} 
\end{comment}
and converts it into training data by picking only the specified features and labels. \Bruce{Features that can be labeled include distance to goal, direction to goal, global obstacle map, local view, agent position, etc.}

\begin{comment} % Sajad 6000

%The training data is then saved as a \texttt{.pickle} file with name format as \texttt{training\_\{atlas name\}\_\{number of samples\}}. The structure of the training data is based on normal \textit{python} objects (\texttt{List[Dict[str, Any]]}) for quick inspection and analysis. 
Features/labels %are picked by using the {MapProcessing} component. These data 
include agent and goal positions, global map, local view, valid moves, etc.  
%(See Table \ref{tab: gen_label_list} for feature reference). 
A* is used as ground truth for feature/label generation. All features/labels can be saved as a variable sequence (needed for LSTM) or single global input (needed for auto-encoder).
\end{comment}

{\it 3) Augmentation.} The augmentation procedure takes an existing training data file and augments it with the specified extra features and labels. It is used to remove the need for re-generating a whole training set.

{\it 4) Modification.} A custom lambda function which takes as input a {map} and returns another {map} and can be defined to modify the underlining structure of the map (e.g. modify the agent position, the goal position, create doors, etc.).

%%%%%%%%%%%%%%%%%%%%%%%%%%%%%%%%%%%%%%%%%%%%%%%%%%%%%%%%%%%%%
%%%%%%% Figure
%%%%%%%%%%%%%%%%%%%%%%%%%%%%%%%%%%%%%%%%%%%%%%%%%%%%%%%%%%%%%
%In the rest of the section, we describe the PathBench platform and its implementation.
%An overview of the PathBench platform design is given in Fig. \ref{fig:sim_architecture}. 

%%%%%%%%%%%%%%%%%%%%%%%%%%%%%%%%%%%%%%%%%%%%%%%%%%%%%%%%%%%%%
%%%%%%% Trainer
%%%%%%%%%%%%%%%%%%%%%%%%%%%%%%%%%%%%%%%%%%%%%%%%%%%%%%%%%%%%%
\subsection{Trainer}

\begin{comment} %% Sajad: removed for the 6000 word limit.
The training pipeline is composed of: (1) data pre-processing, (2) data splitting, (3) training, (4) evaluation, (5) results display, and (6) pipeline end, explained below briefly.
\end{comment}

The training pipeline is composed of six modules, explained below briefly.

{\it 1) Data Pre-processing.} Data is loaded from the specified training sets, and only the features and labels used throughout the model are picked from the training set and converted to a PyTorch {dataset}.

{\it 2) Data Splitting.} The pre-processed data is shuffled and split into three categories: \Bruce{training, validation and testing (Default data split is 60\%, 20\%, and 20\%), according to the holdout method \cite{Goodfellow_2016,Nielsen_2015}}. The {CombinedSubsets} object is used to couple the feature dataset and label dataset of the same category into a single dataset. Then, all data is wrapped into its {DataLoader} object with the same batch size as the training configuration \Bruce{(Default size is 50). These parameters are configurable.}

{\it 3) Training.} The training process puts the model into training mode and takes the training {DataLoader} and validation {DataLoader} and feeds them through the model $n$ times, where $n$ is the number of specified epochs.

\begin{comment} %% Sajad: removed for the 6000 word limit.
The training mode allows the gradients to be updated and at each new epoch, the optimizer sets all gradients to 0. 

Each model has to extend a special \texttt{batch\_start} hook function which is called on each new batch. The \texttt{batch\_start} function is responsible for passing the data through the network and returning the loss result. The trainer takes the loss result and applies a backward pass by calling the \texttt{.backward()} method from the loss. Afterwards, the optimizer is stepped, and the weights of the model are updated. The statistics, such as the loss over time, for the training and validation sets are logged by two {EvaluationResults} objects (one for training and one for validation) which are returned to the pipeline. The {EvaluationResults} class contains several hook functions which are called through the training process at their appropriate times: \texttt{start}, \texttt{epoch\_start}, \texttt{epoch\_finish}, \texttt{batch\_start}, \texttt{batch\_finish}, \texttt{finish}. At each epoch end, the {EvaluationResults} object prints the latest results.
\end{comment}

{\it 4) Evaluation.} The evaluation process puts the model into evaluation mode and has a similar structure to the training process. The evaluation mode does not allow gradients to update. 

\begin{comment} % Sajad: removed for the 6000 word limit
The testing dataset is passed only once through the model and an {EvaluationResults} object containing the final model statistics is returned to the pipeline.
\end{comment}

{\it 5) Results Display.} This procedure displays the final results from the three {EvaluationResults} objects (training, validation, testing) and final statistics such as the model loss are printed.

\begin{comment} % Sajad: removed for the 6000 word limit
The training and validation loss logs are displayed as a \textit{matplotlib} \cite{Hunter:2007} figure. This method can be easily extended to provide more insight into the network architecture (e.g. the convolutional autoencoder {(CAE)} model displays a plot which contains the original image, the reconstructed version, the latent space snapshot and the resulting feature maps).
\end{comment}

%\subsubsection*{{Pipeline End}} 
{\it 6) Pipeline End.} At the end, the model is saved by serialising the model \texttt{.state\_dict()}, model configuration, plots from results display process, and full printing log. % into a {ModelSudir} under {ModelDir}. %The save name is formated according to the following convention: \texttt{\{config save\_name\}\_\{config training\_data\}\_model}.

%%%%%%%%%%%%%%%%%%%%%%%%%%%%%%%%%%%%%%%%%%%%%%%%%%%%%%%%%%%%%
%%%%%%% Analyzer
%%%%%%%%%%%%%%%%%%%%%%%%%%%%%%%%%%%%%%%%%%%%%%%%%%%%%%%%%%%%%
\subsection{Analyzer}\label{sec:analyzer}

The {analyzer} is used to assess and compare the performance of the path planners. 
\begin{comment} %Sajad: removed for the 6000 word limit
This is achieved by making use of the {BasicTesting} component. When a new session is run through the {AlgorithmRunner}, statistical measures depending on the type of testing can be collected by attaching a {BasicTesting} component.
%(See Table \ref{tab: a_testing_tab}). 
The {BasicTesting} component is also linked to the simulator to enable visualization testing. The key frame feature and synchronization variable are tied to the {BasicTesting} component, which allows the user to enhance each key frame and define custom behaviour. Each {algorithm} instance can create debugging views called {MapDisplay}s which can render custom information on the screen such as the the internal state of the {algorithm} (e.g. search space, total fringe, graph, map and its entities etc).% (See Table \ref{tab: map_dsiplays}).
\end{comment}
%Instead of manually running a \textbf{Simulator} instance to assess an \textbf{Algorithm}, the \textbf{analyzer} has an extensive algorithmic analysis procedure split into two parts: simple analysis and complex analysis. We also provide a training dataset analysis routine for inspecting the generated maps.
In addition to manually running a {simulator} instance to assess an {algorithm}, the {analyzer} supports the following analysis procedures:

\begin{itemize}
\addtolength{\itemindent}{-.4cm}
\item Simple Analysis. $n$ map samples are picked from each generated map type, and $m$ algorithms are assessed on them. The results are averaged and printed. Barplots and violinplots, for metrics discussed in Sec.~\ref{sec:performance metrics}, are generated with results from Simple Analysis.
\item Complex Analysis. $n$ maps are selected (generated or hand-made), and $m$ algorithms are run on each map $x$ (\Bruce{Default is 50 runs for each map}) times with random agent and goal positions.  %As in the simple analysis stage, 
In the end, all $n \times x$ results are averaged and reported. Similarly to Simple Analysis, barplots and violinplots for selected metrics can be generated with results.
\item Training Dataset Analysis. A training set analyzer procedure is provided to inspect the training datasets by using the basic metrics 
%defined in Table \ref{tab: a_testing_tab} 
(e.g. Euclidean distance, success rate, map obstacle ratio, search space, total fringe, steps, etc. See the website of the project for all metrics and statistics).   
\end{itemize}

%All printing from the three sections is saved in log files in the \textbf{Resources} directory. In order to view and interpret the results in a friendlier format, the results are tabulated. 
%===============================================================================

%%%%%%%%%%%%%%%%%%%%%%%%%%%%%%%%%%%%%%%%%%%%%%%%%%%%%%%%%%%%%
%%%%%%% Supported Algorithms
%%%%%%%%%%%%%%%%%%%%%%%%%%%%%%%%%%%%%%%%%%%%%%%%%%%%%%%%%%%%%
\section{{Supported Path Planning Algorithms}}
\label{sec:supported-algorithms}

\begin{comment} % Sajad 6000

With the advantage of supporting both classical and machine learning based path planning algorithms, PathBench provides a lightweight framework where development and evaluation of algorithms can be conducted. Currently supported algorithms are categorized and introduced in the following and additional algorithms can be implemented to PathBench easily.    

\subsection{{Classical Algorithms}}
\end{comment}

\Sajad{The planning algorithms implemented into PathBench include the following algorithms and categories:} 
\begin{itemize}
    \item \Sajad{Graph-based planners such as A*, Dijkstra, CGDS~\cite{LCPC}, and wavefront;}
    \item \Sajad{Discrete sampling-based algorithms such as simple probabilistic roadmap (d-sPRM) and d-RRT with its variations such as d-RT, d-RRT*, and d-RRT-Connect;} Moreover, additional sampling based algorithms from the Open Motion Planning Library (OMPL) are also added to improve benchmarking capability of PathBench; 
    \item Sensory-based planning algorithms, e.g. Bug1 and Bug2~\cite{sensor1997};
    \item Numerical optimization methods such as potential field algorithm~\cite{pot1992};
    \item Learning-based path planning algorithms including 
Value Iteration Networks (VIN)~\cite{tamar2016value}, 
Motion Planning Networks (MPNet)~\cite{qureshi2019motion}, 
Gated Path Planning Networks (GPPN)~\cite{lee2018gated}, 
Online LSTM~\cite{nicola2018lstm}, 
CAE-LSTM~\cite{inoue2019robot},
Bagging LSTM~\cite{wpnconf}, and
Waypoint Planning Networks~(WPN)~\cite{wpnconf}. 
\end{itemize}
 
Additional classical or learning-based algorithms can be implemented to PathBench easily. The algorithms developed inside PathBench support step-by-step planning animation, which is an interesting feature for debugging and educational purposes.

\Sajad{We have included sampling-based planners, although they were originally designed for planning continuous space, and have competitive results in such settings. Graph-based or discrete sampling-based algorithms also exist, where instead of sampling from the free space, nodes of the graph are sampled for planning.} 
%\Bruce{Sampling-based planners were originally designed for planning in continuous space, and they have competitive results in such settings.} 
%\Sajad{Graph-based or discrete sampling-based algorithms also exist, where instead of sampling from the free space, nodes of the graph are sampled for planning.}
\Sajad{Examples of such algorithms include \cite{Branicky_ICDC_2003,Morgan_IROS_2004,Hvezda_conf_2018}, d-RRT \cite{Solovey_IJRR_2016}, and d-RRT*~\cite{Shome_AR_2020}; however, as Branicky et. al state \cite{Branicky_ICDC_2003}, ``the performance of the RRT in discrete space is degraded by a decreased bias toward unexplored areas.''} \Sajad{For benchmarking purposes, discretized versions of sampling algorithms (e.g. discretized RRT-Connect and discretized sPRM) were implemented within PathBench where each random sample is confined to an available grid cell. %to ensure fair comparisons~\cite{Morgan_IROS_2004}.
}  

\begin{figure}
  \begin{minipage}[b]{0.32\columnwidth}
  \centering
    \includegraphics[width=.99\columnwidth]{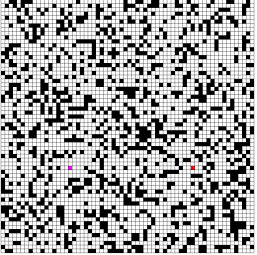}\\
    \includegraphics[width=.95\columnwidth]{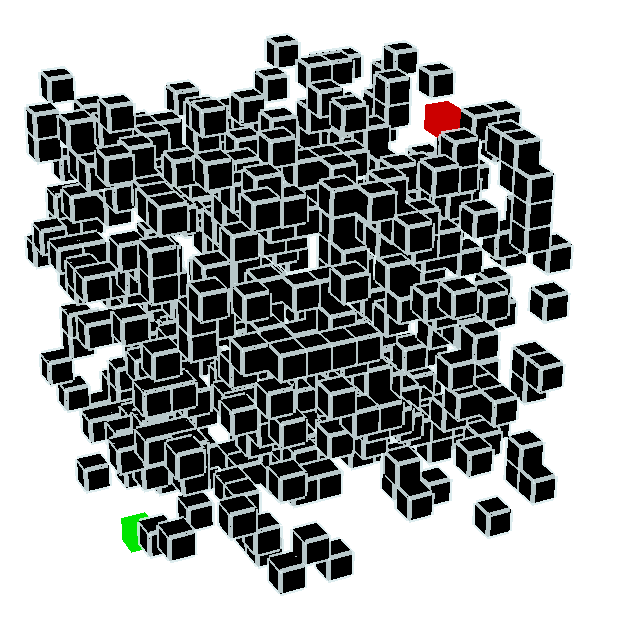} \\
    (a) 
  \end{minipage}
  \begin{minipage}[b]{0.32\columnwidth}
  \centering
    \includegraphics[width=.99\columnwidth]{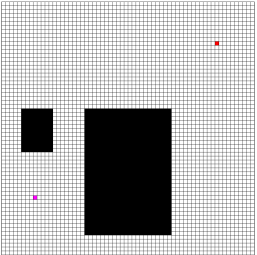}\\
    \includegraphics[width=.99\columnwidth]{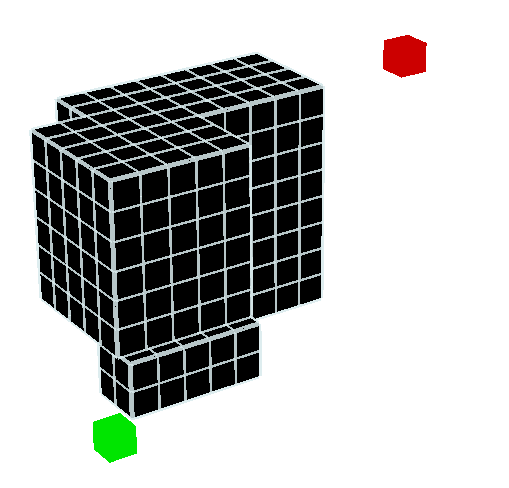} \\
    (b)
  \end{minipage}
  \begin{minipage}[b]{0.32\columnwidth}
  \centering
    \includegraphics[width=0.99\columnwidth]{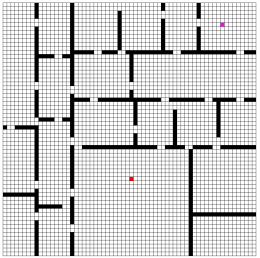}\\
    \includegraphics[width=.99\columnwidth]{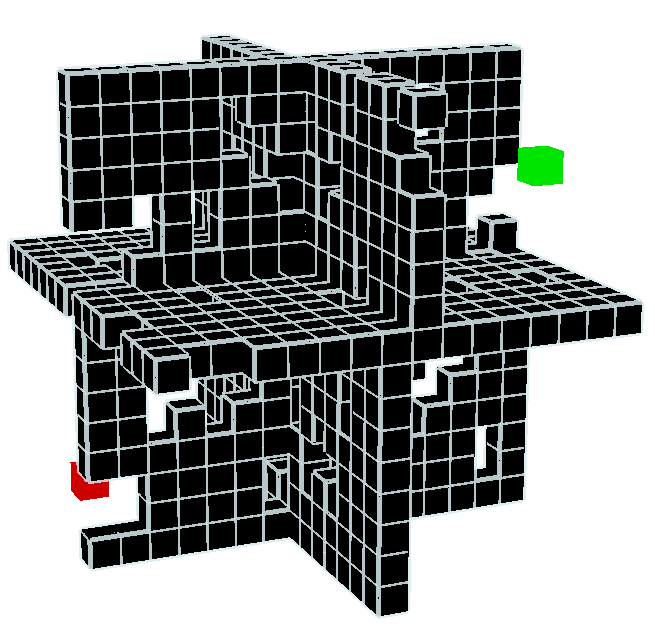} \\
    (c)
  \end{minipage}
  \vspace{1mm}
\caption{The 3 types of generatable grid maps are: (a) Uniform random fill ($64\times64$ 2D and $16\times16$ 3D dimensions, [0.1, 0.3] obstacle fill rate range), (b) Block
map ($64\times64$ 2D and $16\times16$ 3D dimensions, [0.1, 0.3] obstacle fill rate range, [1, 6] number of obstacles range), (c) House atlas ($64\times64$ 2D and $16\times16$ 3D dimensions, [8, 15] minimum room size range, [35, 45] maximum room size range). (Note: Magenta colour is used as goal for all 2D maps as the green goal is difficult to spot.)}
  \label{fig:3maps}  
  \vspace{-2 mm}
  \end{figure}

%%%%%%%%%%%%%%%%%%%%%%%%%%%%%%%%%%%%%%%%%%%%%%%%%%%%%%%%%%%%%
%%%%%%% Supported Maps
%%%%%%%%%%%%%%%%%%%%%%%%%%%%%%%%%%%%%%%%%%%%%%%%%%%%%%%%%%%%%
\section{{Supported Maps}}
\label{sec:maps}
The map is the environment of which simulation and benchmarking of algorithms are performed. \Bruce{Grid environments in 2D and 3D are presently used within PathBench.} A Map contains different entities such as the agent, goal and obstacles, and provides a clean interface that defines the movement and interaction between them. Therefore, a map can be extended to support various environments. The following are supported 2D and 3D map types currently in PathBench.

\subsection{Synthetic Maps}
As mentioned previously in the Generator (Sec.~\ref{sec: generator}), four synthetic map types can be created and used inside PathBench. The simplest map type, block maps, contains a random number of randomly sized blocks that act as obstacles. On the other hand, the map type of uniform random fill maps consists of single obstacles placed at random in the maps' free spaces. The third map type, house maps, aims to mimic typical floorplans by placing obstacles in the form of randomly sized and partitioned walls. Lastly, 3D point cloud maps that contain a set of obstacles in an unbounded 3D space can also be generated and used in PathBench. The inclusion of point cloud maps is to facilitate the development and support of algorithms that work exclusively with point clouds, such as MPNet \cite{qureshi2019motion}. Different map types are included in PathBench, so that map-type specific performance of path planning algorithms can be analyzed further (See Fig.~\ref{fig:3maps}.)

\subsection{Real Maps}
Real-world maps can be utilized inside PathBench with the RosMap class. The RosMap extends 2D occupancy grid maps to integrate the gmapping~\cite{gmapping} and other similar 2D SLAM algorithms by converting the SLAM output image into an internal map environment. The RosMap environment has support for live updates, meaning that algorithms can query an updated view by running a SLAM scan. The map uses simple callback functions to
make SLAM update requests and convert movement actions into network messages using the ROS
publisher-subscriber communication system.

\subsection{External Maps}
\begin{figure}
\centering
    \includegraphics[width=.24\columnwidth]{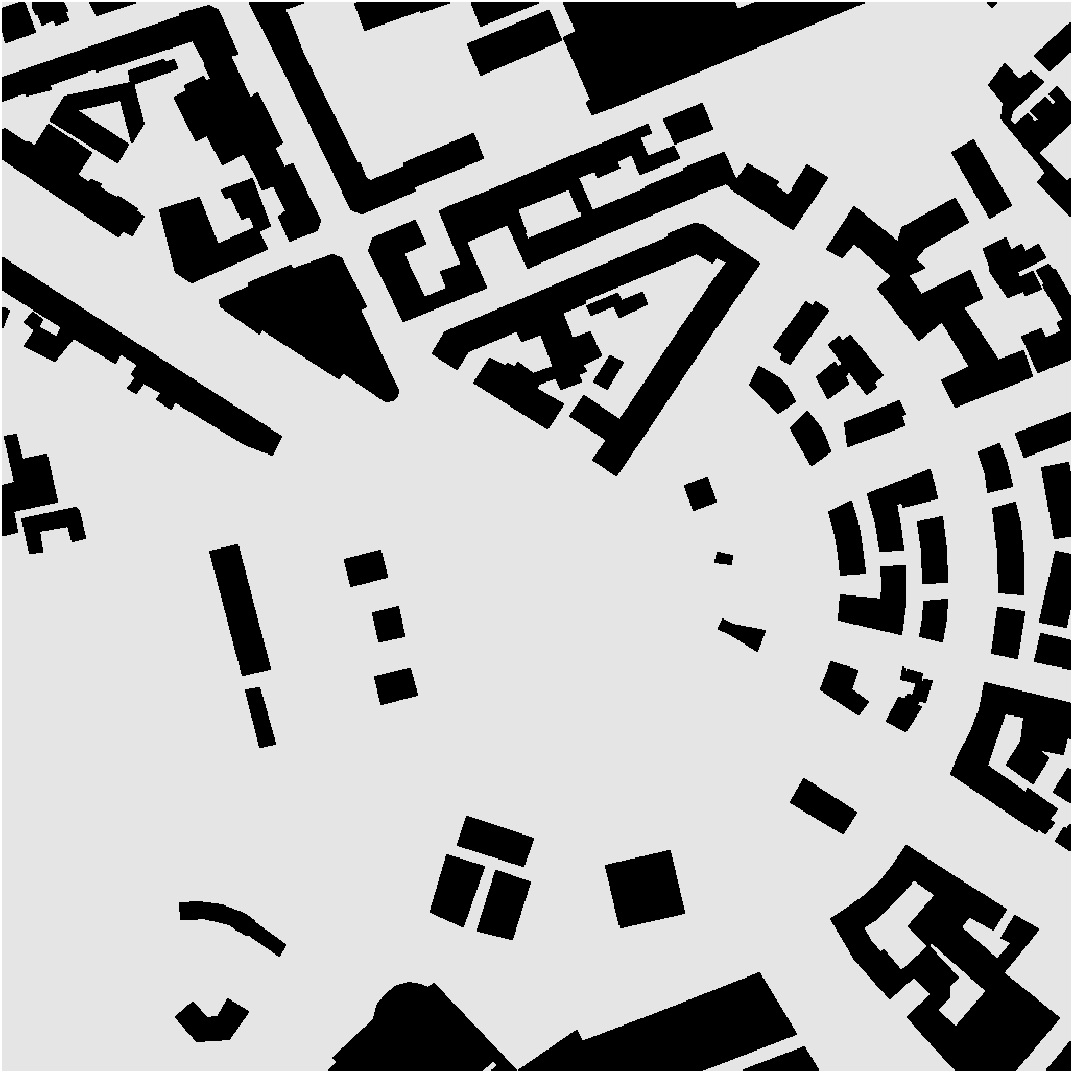}
    \includegraphics[width=.24\columnwidth]{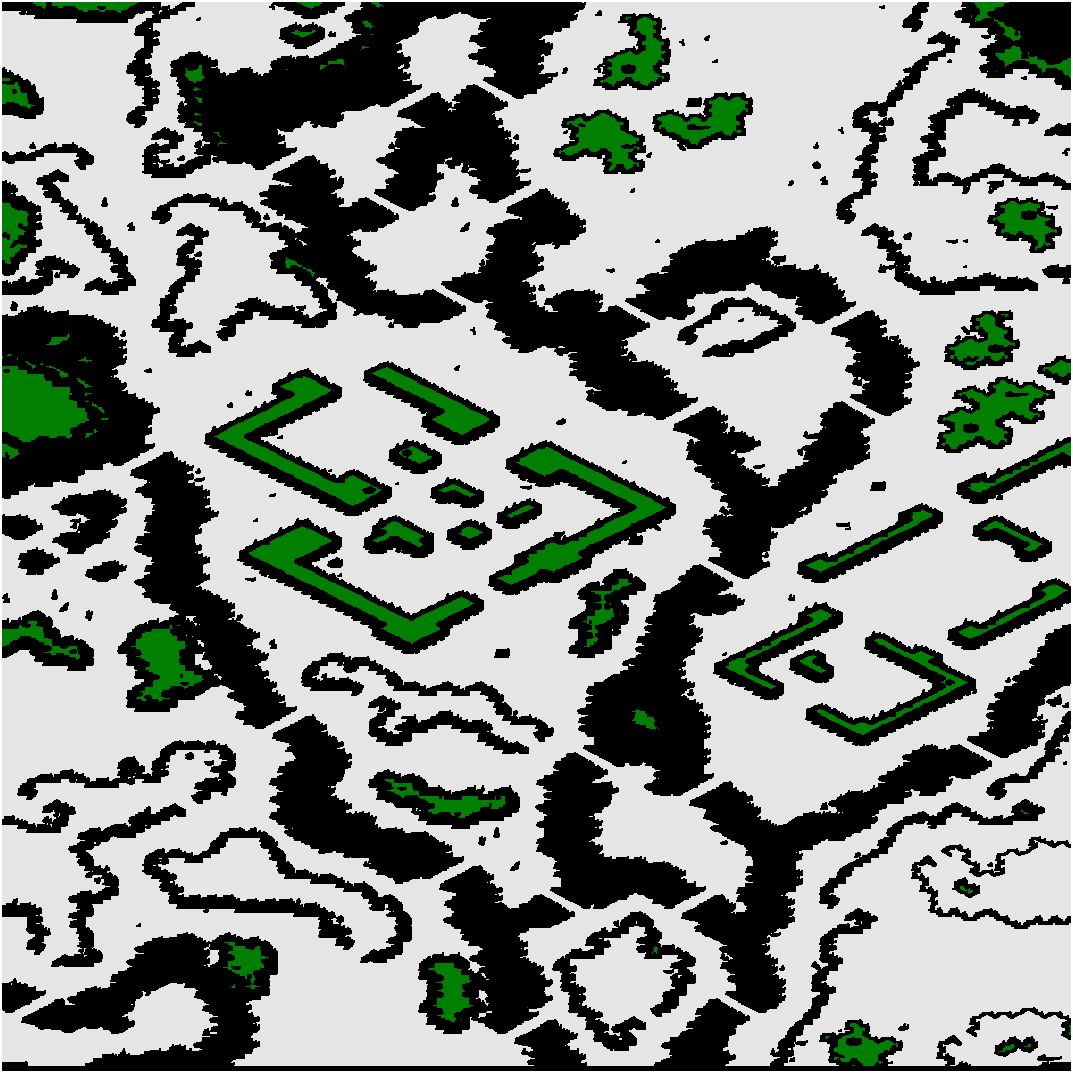}
    \includegraphics[width=.24\columnwidth]{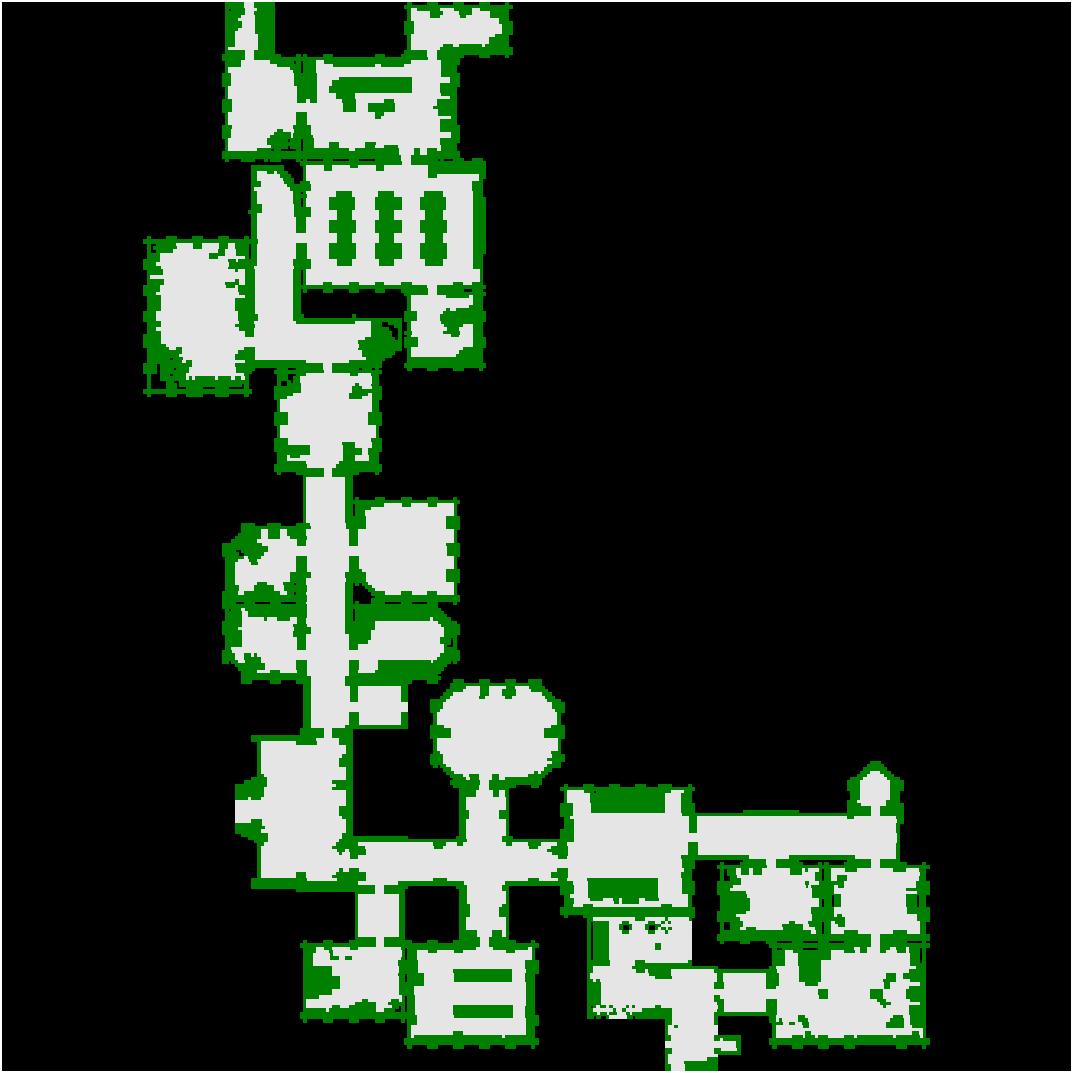}
  \vspace{2mm}
\caption{External maps can be ported into PathBench for benchmarking with ease. The maps above are from video games and real-world city \cite{sturtevant2012benchmarks}.}
  %\label{fig: wp_vs_a_1}
  \label{fig:extmaps}  
  \vspace{-2 mm}
  \end{figure}

External maps can be imported into PathBench to diversify the datasets. Houseexpo~\cite{houseexpo} is a large dataset of 2D floor plans built on SUNCG dataset \cite{song2016ssc}. It contains 35,126 2D floor plans that have 252,550 rooms in total and can be used for PathBench benchmarking. In addition, other video game and real-world datasets can also be converted for PathBench use easily. 2D grid world and 3D voxel maps from video games, such as Warcraft III, Dragon Age and Warframe, and real world \Bruce{discretized 2D grid street maps} from OpenStreetMaps geo-spatial database are implemented into PathBench to demonstrate the ease of integrating external datasets \cite{sturtevant2012benchmarks,brewer2018voxels}, see Fig. \ref{fig:extmaps}. Benchmarking results on external maps are shown in Sec. \ref{sec:result}.

%%%%%%%%%%%%%%%%%%%%%%%%%%%%%%%%%%%%%%%%%%%%%%%%%%%%%%%%%%%%%
%%%%%%% Performance Metrics
%%%%%%%%%%%%%%%%%%%%%%%%%%%%%%%%%%%%%%%%%%%%%%%%%%%%%%%%%%%%%
\section{{Performance Metrics}}
\label{sec:performance metrics}

In order to evaluate and benchmark performance of various algorithms inside PathBench, several metrics are chosen, including success rate, path length, distance left to goal when failed, time, path deviation, search space, memory consumption, obstacle clearance and smoothness of trajectory. Algorithm selection can be aided by evaluating the benchmarked results of task-specific metrics. The following outlines the metrics and rationales behind their selection.

{\it 1) Success Rate (\%).} The rate of success of finding a path, from start to goal, demonstrates the reliability of the algorithm.  

{\it 2) Path Length (cell count).} The total distance taken to reach the goal showcases the efficiency of the path generated.   

{\it 3) Distance Left To Goal (cell count).} The Euclidean distance left
from the agent to the goal, in case of a algorithm failure. This shows the extent of the planning failure.  

{\it 4) Time (seconds).} The total time taken to reach the goal. Time required for planning is an important factor for real life robotics applications. 

{\it 5) Path Deviation (\%).} The path length difference \Bruce{when compared to the shortest possible path, generated by A*.}  

{\it 6) Search Space (\%).} The amount of space that was explored and used to find the final path to the goal. 

{\it 7) Maximum Memory Consumption (MB).} The maximum amount of memory used during a path generation session. Memory usage could be a limiting factor for various robotics settings, thus being a relevant benchmarking metric. 

{\it 8) Obstacle clearance (cell count).} Obstacle clearance provides the mean distance of the agent from obstacles during traversal, \textcolor{black}{where a higher value is typically more ideal.} 

{\it 9) Smoothness of trajectory (degrees).} The average angle change between consecutive segments of paths shows how drastic and sudden the agent's movement changes could be. \Bruce{A lower average angle change value provides a smoother trajectory. With discretized grid environments, smoothness of trajectory could produce misleading values due to large angle changes between each step. Smoothness of trajectory would be a more reliable metric in continuous map types where consecutive path segments can be defined with more clarity. PathBench has plans to explore continuous environments for learned manipulator planning approaches in the future.}  

Other than the metrics above, additional metrics can be implemented into PathBench if required. Nowak {\it et al.} provide potential metrics that could be added, including orientation error, number of collisions, number of narrow passages traversed and number of parameters to tune \cite{nowak2010}.

%%%%%%%%%%%%%%%%%%%%%%%%%%%%%%%%%%%%%%%%%%%%%%%%%%%%%%%%%%%%%
%%%%%%% Experimental Results
%%%%%%%%%%%%%%%%%%%%%%%%%%%%%%%%%%%%%%%%%%%%%%%%%%%%%%%%%%%%%
\section{Experimental Results}
\label{sec:result}

% In this section, experimental results are presented. The experiments include: path planning in simulated and real-world occupancy grid maps, and path planning and navigation using a real robot in real-world office environment. The training of the algorithm is performed using simulated maps only.

% \subsection{Simulated Environment}
% In this section, first the simulated dataset is briefly explained, then the metrics to evaluate the algorithm are described.

% \paragraph{Dataset.}
% For training, three types of synthetic map of size 64 $\times$ 64 pixels were procedurally generated: uniform random fill map, block map, and house map. The map generation process with sample maps are explained in the Supplementary Materials. %Sample maps are shown in Figure \ref{fig: eval_generated maps}. %The analysis of the training datasets can be found in Table \ref{tab: eval_maps}.
% Fig.~\ref{fig: Way runs comp} shows samples of these maps. In these maps, start and goals point are chosen randomly. Evaluations are done over maps that has never been seen by the algorithm. 

% \paragraph{Evaluation Metrics.} To evaluate the generated maps four metrics are used: success rate, trajectory length, distance left to goal when failed, and session search space. To demonstrate why such an architecture was chosen for WPN, we compare the proposed algorithm not only with A*, but also with other modules described in the paper, mainly \emph{Online LSTM}, \emph{CAE-LSTM} and \emph{bagging module}.

In this section, several experiments, \textcolor{black}{including algorithmic benchmarking and hardware system benchmarking}, using PathBench, with classical and learned planners on different maps are presented. \textcolor{black}{The algorithms are evaluated on 2D and 3D synthetic maps of varying sizes that are generated inside PathBench. In addition, video game and street maps from external datasets \cite{sturtevant2012benchmarks} are used for further algorithm benchmarking. Finally, algorithms are tested in ROS and Gazebo with PathBench's ROS extension to highlight its ability to integrate with real world robotics applications.} 

\subsection{Algorithmic Benchmarking}
To begin, classical and learned algorithms, currently supported by PathBench, are benchmarked inside PathBench with different maps. All results are produced by PathBench on Ubuntu 18.04 with an Asus laptop with Intel Core i5-6200U CPU and Nvidia GeForce 940MX. This computer was chosen due to its GPU and processing power being widely available. For training of the learned algorithms, 
%WPN, VIN, MPNet and GPPN, 
three types of synthetic map of size $64 \times 64$ pixels were procedurally generated: uniform random fill map, block map, and house map. 
%
%The map generation process with sample maps are explained in the Supplementary Materials. 
%
%Sample maps are shown in Figure \ref{fig: eval_generated maps}. %The analysis of the training datasets can be found in Table \ref{tab: eval_maps}.
Fig.~\ref{fig:3maps} shows samples of these maps. In these maps, start and goal points are chosen randomly. Evaluations are done on maps that have never been seen by the algorithms. \Bruce{To ensure that comparisons were done with the same trajectories between systems, the pseudorandom number generator's random seed for sampling-based algorithms were initialized identically.} \Sajad{We use the default values for the holdout method since there is a lot of training data available, but if one wants to change the amount of data for training and validation, it is also possible. The default batch size is 50, as it would fit on most GPUs.} 
%
%\textbf{Evaluation Metrics.} 
%To evaluate the generated maps, nine metrics are used: success rate, path length, distance left to %goal when failed, time, path deviation, search space, obstacle clearnace, smoothness of trajectory %and memory consumption.

%\begin{figure}
%  \hfill
%  \begin{minipage}[b]{0.16\textwidth}
%  \centering
%    \vspace{-2 mm}
%    \includegraphics[width=\linewidth]{images/m1.png}
%  \end{minipage}
%  \hfill
%  \begin{minipage}[b]{0.16\textwidth}
%  \centering
%    \vspace{-2 mm}
%    \includegraphics[width=\linewidth]{images/m2.png}
%  \end{minipage}
%  \hfill
%  \begin{minipage}[b]{0.16\textwidth}
%  \centering
%    \vspace{-2 mm}
%    \includegraphics[width=\linewidth]{images/m3.png}
%  \end{minipage}
%  \hfill
%  \begin{minipage}[b]{0.16\textwidth}
%  \centering
%    \vspace{-2 mm}
%    \includegraphics[width=\linewidth]{images/m4.png}
%  \end{minipage}
%  \hfill
%  \begin{minipage}[b]{0.16\textwidth}
%  \centering
%    \vspace{-2 mm}
%    \includegraphics[width=\linewidth]{images/m5.png}
%  \end{minipage}
%  \hfill
%  \begin{minipage}[b]{0.16\textwidth}
%  \centering
%    \vspace{-2 mm}
%    \includegraphics[width=\linewidth]{images/m6.png}
%  \end{minipage}
%  \hfill  
%\caption{Synthetic maps used to generate results in Table II.}
%  \label{fig:synthetic_maps}  
%  \end{figure}

\begin{table}[t]
    \centering
    \caption{Results of classical algorithms: On 2D 64$\times$64 PathBench built in maps (3000 samples), 512$\times$512 city maps (300 samples), and video game maps with 800 to 1200 cells in dimension (300 samples). An Asus laptop with Intel i5-6200U CPU and Nvidia GeForce 940MX was used. The failed cases occur when there is no valid path towards the given goal.}
    \vspace{1mm}
    \begin{tabular}{|c|c|c|c|c|c|}
       \hline
       %\textbf{\scriptsize planner} & \makecell {\textbf{\scriptsize success} \\ \textbf{\scriptsize %rate} } & \makecell {\textbf{\scriptsize path len.} \\ \textbf{\scriptsize }{\scriptsize} } & %\makecell {\textbf{\scriptsize distance left} \\ \scriptsize{(when failed)} } &  %\makecell{\textbf{\scriptsize time} \\ (sec)} & \makecell {\textbf{\scriptsize path} \\ %\textbf{\scriptsize deviation} }  \\
    \makecell {\textbf{\scriptsize map} \\ \textbf{\scriptsize type}} & \textbf{\scriptsize planner} & \makecell {\textbf{\scriptsize path} \\ \textbf{\scriptsize dev.}\scriptsize{(\%)}}  & \makecell {\textbf{\scriptsize distance left} \\ (if failed,cell count)} &\makecell{\textbf{\scriptsize time} \\ (sec)} & \makecell {\textbf{\scriptsize success} \\ \textbf{\scriptsize rate} \scriptsize{(\%)} }\\ 
       \hline
      \parbox[t]{2mm}{\multirow{6}{*}{\rotatebox[origin=c]{90}{PathBench}}} 
      &{\scriptsize A* \cite{duchovn2014path}}  & 0.00 & 0.25 & 0.106 & 99.4  \\%& 10.5\% \\
       \cline{2-6}
       &{\scriptsize Wavefront \cite{luo2014effective}}  & 0.36 & 0.25 & 0.340 & 99.4 \\%& -\% \\
       \cline{2-6}
       &{\scriptsize Dijkstra \cite{choset2005principles}}  & 0.00 & 0.25 & 0.338  & 99.4 \\%& 42.7 \\
       \cline{2-6}
       &{\scriptsize \Sajad{d-sPRM} \cite{kavraki1994probabilistic}}   & 35.03 & 2.67 & 0.767 & 93.3 \\%& 42.7 \\
       \cline{2-6}
       &{\scriptsize \Sajad{d-RRT}~\cite{lavalle1998rapidly}} & 19.91 & 0.48 & 7.336 & 96.8 \\%& - \\
       %\cline{2-6}
       %&{\scriptsize RRT* \cite{rrtstar}} & 6.29 & 1.46 & 9.412  & 99.33 \\%& - \\
       \cline{2-6}
       &{\scriptsize \Sajad{d-RRT-Connect}~\cite{rrtconnect}} & 21.17 & 0.26 & 0.2458 & 99.33  \\%& - \\
       \hline
       \hline
       \parbox[t]{2mm}{\multirow{6}{*}{\rotatebox[origin=c]{90}{City}}} 
       &{\scriptsize A*}            & 0.00 & 0.00 & 3.816 & 100.0 \\%& 10.5\% \\
       \cline{2-6}
       &{\scriptsize Wavefront}     & 1.08 & 0.00 &  8.468 & 100.0\\%& -\% \\
       \cline{2-6}
       &{\scriptsize Dijkstra}   & 0.00 & 0.00 & 9.928 & 100.0 \\%& 42.7 \\
       \cline{2-6}
       &{\scriptsize \Sajad{d-sPRM}}   & 123.64 & 13.68 & 5.377 & 93.6  \\%& 42.7 \\
       \cline{2-6}
       &{\scriptsize \Sajad{d-RRT}} & 54.98 & 3.43 & 39.248 & 95.7 \\%& - \\
       %\cline{2-6}
      % &{\scriptsize RRT* } & 32.71 & 4.11 & 43.464  & 95.1 \\%& - \\
       \cline{2-6}
       &{\scriptsize \Sajad{d-RRT-Connect}} & 65.08 & 5.35 & 3.489 & 96.6 \\%& - \\
       \hline
       \hline
       \parbox[t]{2mm}{\multirow{6}{*}{\rotatebox[origin=c]{90}{Video Games}}} 
      &{\scriptsize A*}               & 0.00 & 0.00 & 31.567 & 100.0 \\%& 10.5\% \\
       \cline{2-6}
       &{\scriptsize Wavefront}     & 1.65 & 0.00 & 43.517 & 100.0 \\%& -\% \\
       \cline{2-6}
       &{\scriptsize Dijkstra}   & 0.00 & 0.00 & 42.366 & 100.0 \\%& 42.7 \\
       \cline{2-6}
       &{\scriptsize \Sajad{d-sPRM}}   & 287.21 & 0.00 & 44.498 & 100.0 \\%& 42.7 \\
       \cline{2-6}
       &{\scriptsize \Sajad{d-RRT}} & 128.90 & 64.38 & 64.881 & 42.6 \\%& - \\
       %\cline{2-6}
       %&{\scriptsize RRT* } & 104.30 & 56.21 & 68.971  & 36.7 \\%& - \\
       \cline{2-6}
       &{\scriptsize \Sajad{d-RRT-Connect}} & 112.90 & 25.92 & 30.84 & 95.30 \\%& - \\
       \hline
%       {\scriptsize Iterative  \cite{qureshi2018motion}} & \% &  &  &  &- \\
%       \hline
    \end{tabular}
    \label{tab:classicresult}
        \vspace{-2 mm}
\end{table}

\begin{table}[t]
    \centering
    \caption{Results of learned algorithms: On the same 2D 64$\times$64 PathBench built in maps (3000 samples), 512$\times$512 city maps (300 samples), and video game maps (300 samples) from Table~\ref{tab:classicresult}. An Asus laptop was used to generate results. (Intel i5-6200U CPU and Nvidia GeForce 940MX)}
    \vspace{1mm}
    \begin{tabular}{|c|c|c|c|c|c|}
       \hline
       %\textbf{\scriptsize planner} & \makecell {\textbf{\scriptsize success} \\ \textbf{\scriptsize %rate} } & \makecell {\textbf{\scriptsize path len.} \\ \textbf{\scriptsize }{\scriptsize} } & %\makecell {\textbf{\scriptsize distance left} \\ \scriptsize{(when failed)} } &  %\makecell{\textbf{\scriptsize time} \\ (sec)} & \makecell {\textbf{\scriptsize path} \\ %\textbf{\scriptsize deviation} }  \\
    \makecell {\textbf{\scriptsize map} \\ \textbf{\scriptsize type}} & \textbf{\scriptsize planner} & \makecell {\textbf{\scriptsize path } \\ \textbf{\scriptsize dev.}\scriptsize{(\%)}} & \makecell {\textbf{\scriptsize distance left} \\ (if failed,cell count)} &\makecell{\textbf{\scriptsize time} \\ (sec)} & \makecell {\textbf{\scriptsize success} \\ \textbf{\scriptsize rate}\scriptsize{(\%)}}\\ 
       \hline
      \parbox[t]{2mm}{\multirow{7}{*}{\rotatebox[origin=c]{90}{PathBench}}} 
      &{\scriptsize VIN \cite{tamar2016value}}      &5.25 & 29.55 & 1.796 & 18.2   \\%&  \\
       \cline{2-6}
       &{\scriptsize MPNet \cite{qureshi2019motion} } & 10.86 &  21.59 &  0.296 & 41.2  \\%& - \\
       \cline{2-6}
       &{\scriptsize GPPN \cite{lee2018gated}} & 6.83 & 27.56 & 2.018 & 30.3  \\%& - \\
       \cline{2-6}
       &{\scriptsize Online LSTM~\cite{nicola2018lstm} }   & 0.79 & 11.23 & 0.162 & 53.8\\%& 42.7 \\
       \cline{2-6}
       &{\scriptsize CAE-LSTM~\cite{inoue2019robot}}   & 0.99 & 12.57 & 0.207 & 48.7\\%& 42.7 \\
       \cline{2-6}
       &{\scriptsize Bagging LSTM \cite{wpnconf}} & 1.62 & 3.99 & 0.985 & 78.1 \\%& - \\
       \cline{2-6}
       &{\scriptsize WPN \cite{wpnconf}} & 2.78 & 0.15 & 0.817 & 99.4 \\%& -\% \\
       \hline
       \hline
       \parbox[t]{2mm}{\multirow{7}{*}{\rotatebox[origin=c]{90}{City}}} 
      &{\scriptsize VIN}               & 100.00 &184.21 & 32.612 & 0.0 \\%& 10.5\% \\
       \cline{2-6}
       &{\scriptsize GPPN } & 100.00 & 184.21 &  57.633 & 0.0 \\%& - \\     
       \cline{2-6}
       &{\scriptsize Online LSTM}   & 1.05 &  87.51 & 3.328 & 16.7\\%& 42.7 \\
       \cline{2-6}
       &{\scriptsize CAE-LSTM}   & 4.49 & 91.20 & 4.241 & 10.0\\%& 42.7 \\
       \cline{2-6}
       &{\scriptsize Bagging LSTM} & 31.67 &  45.06 & 20.268 & 43.3  \\%& - \\
       \cline{2-6}
       &{\scriptsize WPN}             & 8.44 & 0.00 & 13.816  & 100.0 \\%& -\% \\
       \hline
       \hline
       \parbox[t]{2mm}{\multirow{7}{*}{\rotatebox[origin=c]{90}{Video Game}}} 
      &{\scriptsize VIN}               & 100.00 & 276.30 & 42.19 & 0.0 \\%& 10.5\% \\
       \cline{2-6}
       &{\scriptsize GPPN } & 100.00 & 276.30 & 65.877 & 0.0 \\%& - \\
       \cline{2-6}
       &{\scriptsize Online LSTM}   & 0.00 & 217.10 & 16.380 & 8.3 \\%& 42.7 \\
       \cline{2-6}
       &{\scriptsize CAE-LSTM}   & 3.05 & 199.60 & 27.330 & 5.3\\%& 42.7 \\
       \cline{2-6}
       &{\scriptsize Bagging LSTM } & 0.41 & 155.89 & 123.901 & 20.6  \\%& - \\
       \cline{2-6}
       &{\scriptsize WPN}             & 10.55 & 0.00  & 110.307  & 100.0 \\%& -\% \\
       \hline
%       {\scriptsize Iterative  \cite{qureshi2018motion}} & \% &  &  &  &- \\
%       \hline
    \end{tabular}
    \label{tab:learnedresult}
        \vspace{-2 mm}
\end{table}

\begin{figure}[t]
\centering
    \includegraphics[scale=0.3]{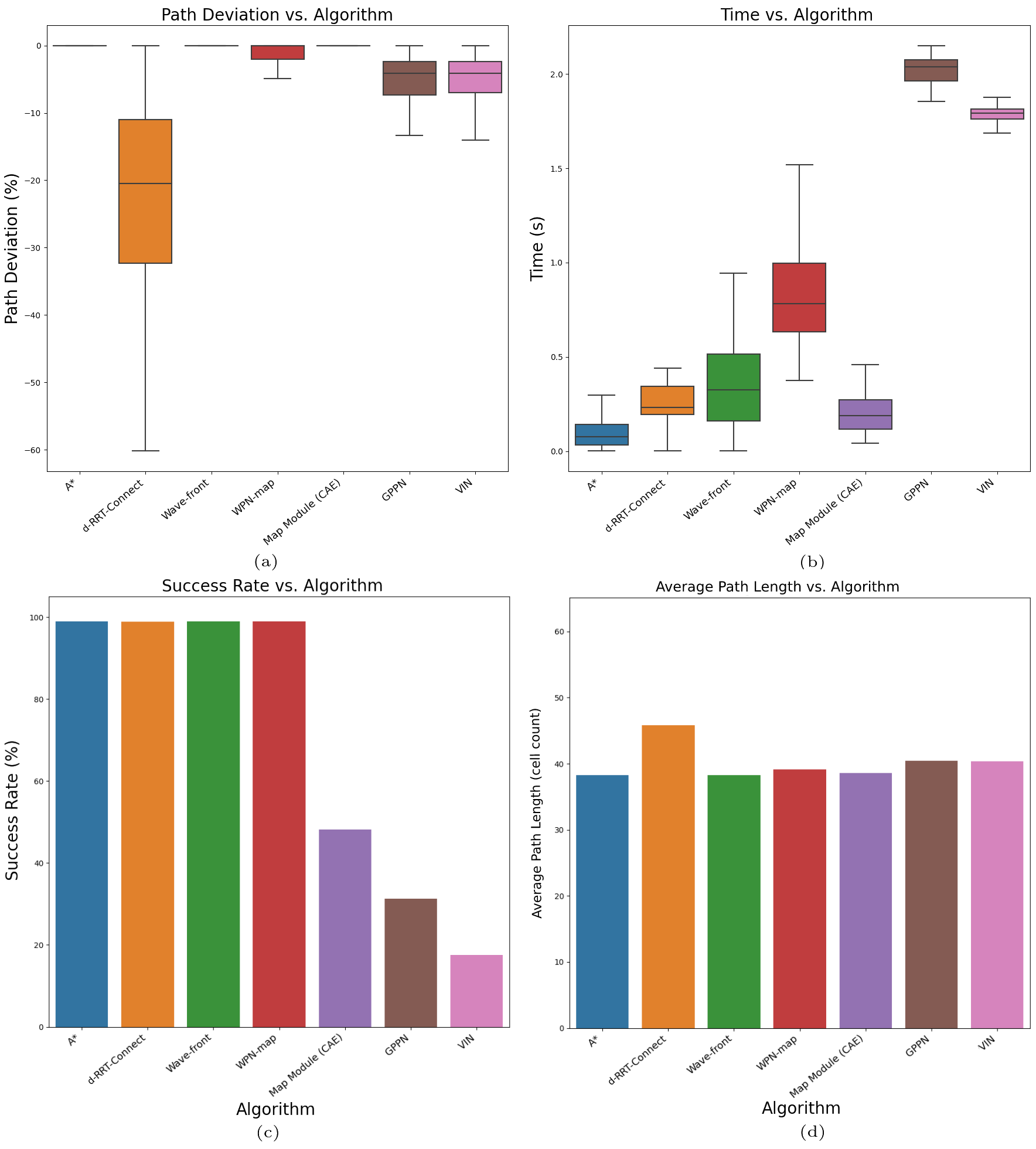}
    %\includegraphics[width=.47\columnwidth]{images/bar2.png}
    %\includegraphics[width=.47\columnwidth]{images/vio1.png}
    %\includegraphics[width=.45\columnwidth]{images/vio2.png}
  %\vspace{-4mm}
\caption{Graphical analysis of 2D benchmarking for classical and learned algorithms \Bruce{ that is produced by PathBench's analyzer}. Results from Table~\ref{tab:classicresult} and Table~\ref{tab:learnedresult} are used. }
  %\label{fig: wp_vs_a_1}
  \label{fig:barvio}  
  \vspace{-2 mm}
  \end{figure}
  
%%%%%%%%%%%%%%%%%%%%%%%%%%%%%%%
%%%%%%%%% Table 3D results
%%%%%%%%%%%%%%%%%%%%%%%%%%%%%%%
%%%%%%%%%%%%%%%%%%%%%%%%%%%%%%%

\begin{table}
    \centering
    \caption{Results of classical algorithms on 3D 28$\times$28$\times$28 PathBench built-in maps using an Asus laptop (3000 samples).}
    \vspace{1mm}
    \begin{tabular}{|c|c|c|c|c|c|}
       \hline
       \textbf{\scriptsize planner} & \makecell {\textbf{\scriptsize success} \\ \textbf{\scriptsize rate}{\scriptsize (\%)} } & \makecell {\textbf{\scriptsize path } \\ {\scriptsize \textbf{len.}(cell count)} } & \makecell {\textbf{\scriptsize path} \\ \textbf{\scriptsize dev. {\scriptsize(\%)}} }&  \makecell{\textbf{\scriptsize time} \\ (sec)} & \makecell {\textbf{\scriptsize path smooth-} \\ \textbf{\scriptsize ness}{\scriptsize (deg.)}}   \\
       \hline
       {\scriptsize A*}               & 100.0 & 20.69 & 0.00 & 0.475 & 0.28 \\
       \hline
       {\scriptsize Wavefront}              & 100.0 & 21.27  &  0.51 & 6.118 & 0.11 \\
       \hline
       {\scriptsize Dijkstra}   & 100.0 & 20.69 & 0.00 & 8.453 & 0.13 \\
       \hline
       {\scriptsize \Sajad{d-sPRM}}   & 100.0 & 36.87 & 16.18 & 0.248 & 0.37 \\
       \hline
       {\scriptsize \Sajad{d-RRT-Connect}} & 99.7 & 38.22 & 17.56 & 0.097 & 0.41 \\
       \hline
%       {\scriptsize Iterative  \cite{qureshi2018motion}} & \% &  &  &  &- \\
%       \hline
    \end{tabular}
    \label{tab:3dresults}
        \vspace{-2 mm}
\end{table}

\subsubsection{2D Synthetic Maps: Simple Analysis}
To demonstrate the benchmarking ability of PathBench and its support for the machine learning algorithms, 
%WPN, Online LSTM, CAE-LSTM, bagging module, VIN, GPPN and MPNet planners
all the algorithms described in Sec.~\ref{sec:supported-algorithms} are analyzed against classical path planning algorithms in $64 \times 64$ 2D PathBench maps. One thousand maps of each of the three types of PathBench maps were used. Table~\ref{tab:classicresult} and Table~\ref{tab:learnedresult} present detailed comparative results for simple analysis of 3000 2D PathBench maps. Fig.~\ref{fig:barvio} displays some of the key results in bar and violin plots.

\subsubsection{2D External Maps: Complex Analysis}
Both classical and learned algorithms were also benchmarked using the analyzer's complex analysis tool, in order to demonstrate the framework's ability to evaluate algorithm performance on specific map types. The analysis was performed on $n=30$ external city maps from OpenStreetMaps' geo-spatial database \cite{sturtevant2012benchmarks}, with 10 random samples collected for averaging of results on each $512 \times 512$ map. \Sajad{Each map was run with $x=50$, selected as a reasonable value for hyperparameter. See Sec.~\ref{sec:analyzer} for parameter definitions.} Thirty video game maps with height and width varying from 800 to 1200 cells were benchmarked in a similar manner. Results of benchmarking on video game and city maps are also listed in Table~\ref{tab:classicresult} and Table~\ref{tab:learnedresult}. The use of external environments in this experiment demonstrates the capability of PathBench to incorporate additional datasets. 

\subsubsection{3D Maps: Simple Analysis}
To demonstrate PathBench's support for 3D path planning, analysis of path planning algorithms on 3D $28 \times 28 \times 28$ PathBench maps was conducted. The benchmarking results that averaged algorithm performance on 1000 maps of each PathBench map type, uniform random fill map, block map, and house map, is shown in Table~\ref{tab:3dresults}.

By looking at the 2D and 3D results, we can quickly assess some strengths and weaknesses of each planning approach. For example, the three graph-based algorithms all find a solution when one exists. \textcolor{black}{A* and Dijkstra algorithms have path deviation of 0 percent, and wavefront also guarantees paths close to shortest possible lengths.} \Sajad{d-RRT-Connect} has much higher success rate than other sampling-based methods, while being considerably faster. The number and type of samples taken is a parameter that can be modified to configure sampling-based algorithms' behaviour in PathBench. \textcolor{black}{Although A* can generate the shortest path length in both 2D and 3D planning scenarios, \Sajad{d-RRT-Connect} is capable of planning at a significantly faster time in 3D and larger 2D environments. Sampling path planning algorithms are also selected often for planning in higher dimensions and are the standard for manipulator motion planning. They will be focused on heavily as PathBench extends to higher dimensional planning.} Machine learning algorithms, on the other hand, experience lower success rates for all map types. \textcolor{black}{The success rates decrease greatly as maps become larger and more complex. For example, }VIN and GPPN have shown to not scale well with the increase in map size and could not successfully provide any paths in the city and video game datasets. \Sajad{The drop in success rate of VIN and GPPN is consistent with what has been reported in the respective papers. Meanwhile, it is notable from the results of the benchmarking that GPPN outperforms VIN, suggesting that recurrent neural networks can help the algorithms to have a better performance.} WPN is an exception with the ability to plan at 100\% success rate for all map types when a solution is available and has a notably lower path deviation. \textcolor{black}{This is due to WPN being a hybrid algorithm that incorporates A* in its way point planning approach. As learning-based algorithms, the three LSTM algorithms and MPNet have faster planning times and success rates than VIN and GPPN. However, machine learning algorithms' path planning times in general are higher when compared to classical approaches, especially as the map size and complexity increases.} MPNet was only tested on PathBench maps, due to constraint of the implementation used. The publicly available version of the network allowed encoding of a limited number of obstacles. \textcolor{black}{It has a faster planning time than VIN, GPPN and WPN, but a higher path deviation.} Performing this kind of simple and rapid analysis is trivial in PathBench.

\begin{figure}
  \centering
   \resizebox*{10cm}{!}{\includegraphics{images/legend_final2.png}}\hspace{3pt}
  \subfloat[]{%
  \centering
   \resizebox*{7cm}{!}{\includegraphics{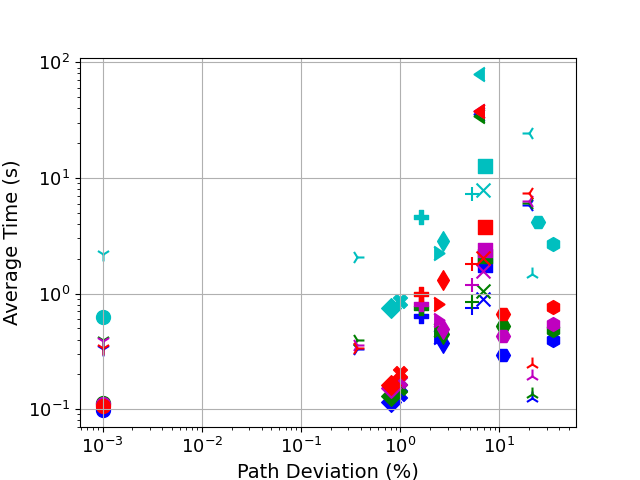}}}\hspace{1pt}
  \subfloat[]{%
   \resizebox*{7cm}{!}{\includegraphics{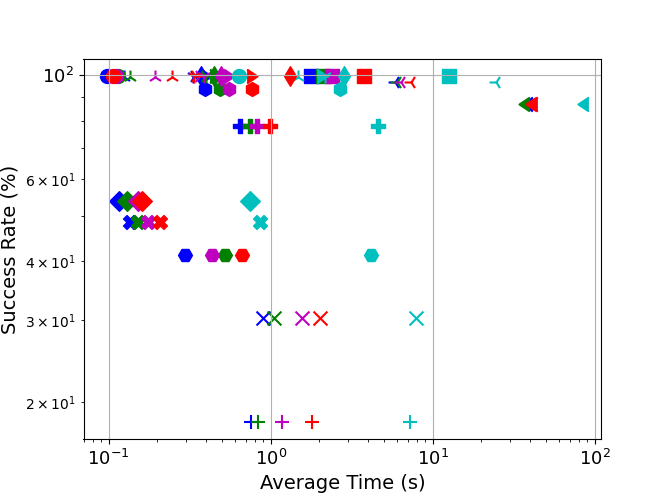}}}\hspace{1pt}
  \subfloat[]{%
   \resizebox*{7cm}{!}{\includegraphics{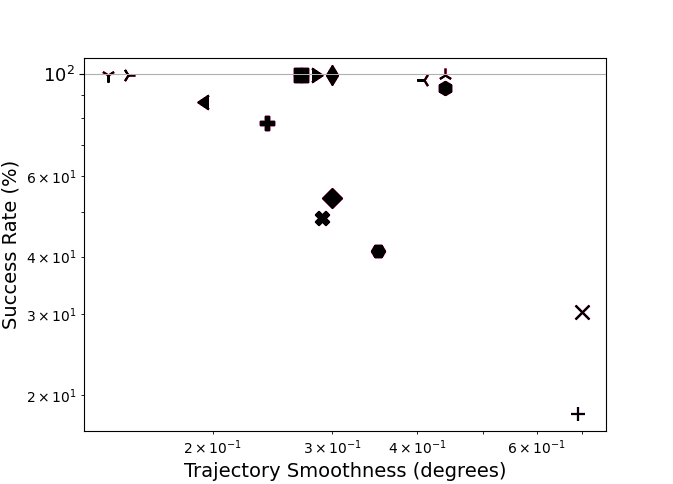}}}\hspace{1pt}
   \subfloat[]{%
   \resizebox*{7cm}{!}{\includegraphics{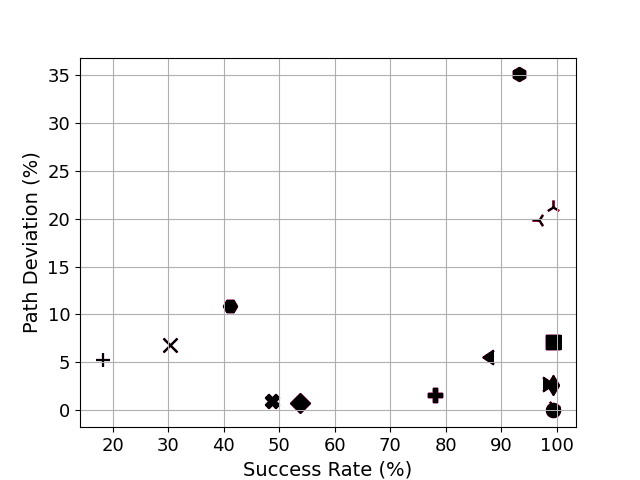}}}\hspace{1pt}
  \vspace{1 mm}
  \caption{\SajadT{Scatter plots showing the performance of several classical and learned path planning algorithms.
Each marker shape shows a different algorithm and different colors demonstrate the various hardwares. Performance metrics include the path deviation, success rate of the algorithm, trajectory smoothness, and the computation time. \textcolor{black}{(Some markers overlap each other completely, due to values being the same across systems. These overlapped markers are indicated in black, as seen in plot (c) and (d).)}}}
  \label{fig:4scatter}
      \vspace{- 3mm}
\end{figure}

\SajadT{
\subsection{Hardware System Benchmarking}
\label{sec:hard}
A comparative study using various hardware systems that are often used in the development and analysis of robotic applications has also been conducted. It demonstrates algorithm performance and consistency of PathBench's benchmarking ability across different systems. Five hardware systems were used as follows:}

\SajadT{{\it 1) Commodity Laptop. } A commonly used laptop is chosen for becnhmarking due to its availability and use for rapid prototyping of path planning algorithms. It is equipped with \Bruce{Intel i5-6200U} CPU and Nvidia GeForce 940MX GPU.}

\SajadT{{\it 2) Gaming Laptop.} The second laptop used possesses more computing power and has an Intel Core i9-10980HK CPU with Nvidia GeForce RTX 2080 GPU.
}

\SajadT{{\it 3) Intel NUC. }The Intel NUC10i7FNK has \Bruce{an Intel i7-10710U CPU with} no GPU and was also used to assess algorithms. It is a small form factor desktop computer and is popular in mobile robotic applications.} 

\SajadT{{\it 4) Compute Canada. }Compute Canada's Graham, a general purpose cluster for a variety of workloads, was utilized. For the Compute Canada analysis, Intel Xeon Gold 5120 CPU and Nvidia V100 GPU were used.} 

\Bruce{{\it 5) Raspberry Pi.} Lastly, a Raspberry Pi Model 4b with ARM Cortex-A72 processor was also analyzed. It has no GPU and has 8GB of RAM. Raspberry Pi is commonly used in mobile robotics applications, making it an important device for path planning analysis.}

\SajadT{The five hardware systems were analyzed on 64$\times$64 2D maps with one thousand maps for each of the three types of PathBench maps, i.e. uniform random fills, block maps, and house maps, shown in Fig.~\ref{fig:3maps} (top row), totalling 3,000 maps.} 

\SajadT{\subsubsection{Benchmarking Analysis}}
\SajadT{By examining Fig. \ref{fig: scatter1} and Fig. \ref{fig:4scatter}, system specific algorithm performances are observed. Each scatter plot displays how optimal the algorithms are for two different metrics. Note that some markers overlap each other completely, as seen in plot Fig. \ref{fig:4scatter}-c and Fig. \ref{fig:4scatter}-d that have mostly black markers. These occur for the deterministic algorithms, such as A* and Dijkstra, in which the variation in hardware do not affect the performance metrics such as success rate or trajectory smoothness.}

\par
\SajadT{The computation times of the algorithms for hardware systems with more powerful CPUs are significantly faster as expected. Systems with more advanced GPUs have lower computation time for learning-based algorithms, such as VIN, GPPN and WPN, due to machine learning algorithms' reliance on GPUs for matrice operations. A consistent spread across data points for different systems in the average time metric is observed. On the other hand, success rate (Fig.~\ref{fig:4scatter}-b and c) and path deviation (Fig.~\ref{fig:4scatter}-a and d) of both classical and learning-based algorithms share a strong consistency across the five hardware systems tested. Trajectory smoothness and obstacle clearance performance of classical and learned algorithms are also consistent across platforms.}

%However, slight variation in success rate, trajectory smoothness and path deviation can be caused by the built-in randomness of sampling-based and some learning-based algorithms. RRT, RRT-Connect and MPNet experience this randomness and Fig. \ref{fig:4scatter}-c and d confirm the hypothesis with non-overlapping markers as the three algorithms do \textcolor{black}{not produce identical trajectory smoothness and path deviation across systems. These algorithms terminate after certain number of tries, if a solution is not found. This slight variation in algorithm performance can be minimized further by increasing the sample size of benchmarking.}}

\par
\SajadT{Specific analysis between algorithms can be done as well. A* and WPN share the same trajectory smoothness in Fig.~\ref{fig:4scatter}-c, as seen with the overlapped black markers. This is due to WPN's use of A* for planning between waypoints. Dijkstra is expected to have similar trajectory with A*; however, the existence of multiple shortest possible paths on most maps led to a different trajectory smoothness value. On the other hand, GPPN was built upon VIN to improve its lower rate of planning success. In Fig. \ref{fig:4scatter}-b, GPPN can be observed to require more time for path computation in PathBench while having a much higher success rate than VIN. However, GPPN's computation time can be improved significantly and surpass VIN's performance on most systems, if \Bruce{CPUs with better processing power are used.} This, combined with consistency of other metrics across systems, suggests that computation time of algorithms is the most important factor when selecting hardware systems for robotic applications.}

\SajadT{Scatter plots seen in this section can be used to benchmark and select optimal algorithms and hardware for specific tasks and can be generated with PathBench's analyzer. Such analysis, enabled by PathBench, allow system developers and algorithm designers to efficiently select the right algorithm for their application or benchmark and compare their new algorithm with existing ones.} \Sajad{For instance, if for a particular application, a higher obstacle clearance is needed, according to Fig.~\ref{fig: scatter1}, compared with A*, MPNet is a better choice assuming there is access to a good GPU such as NVidia GeForce RTX 2080, since MPNet is able to generate paths in $\approx$0.3 seconds with the best obstacle clearance values. If no GPU is available, still the algorithm works under a second, i.e. $\approx$0.5 seconds on NUC. This advantage, according to Fig.~\ref{fig:4scatter}-b, comes at the cost of having less success rate ($\approx$41\%) compared with other machine learning algorithms such as WPN ($\approx$99\%) and Bagging LSTM ($\approx$78\%).}

\newpage
\vspace{5 mm}
\subsection{Real-world Robot Interfacing}
\begin{figure}[t]
\centering
    \includegraphics[scale=0.113]{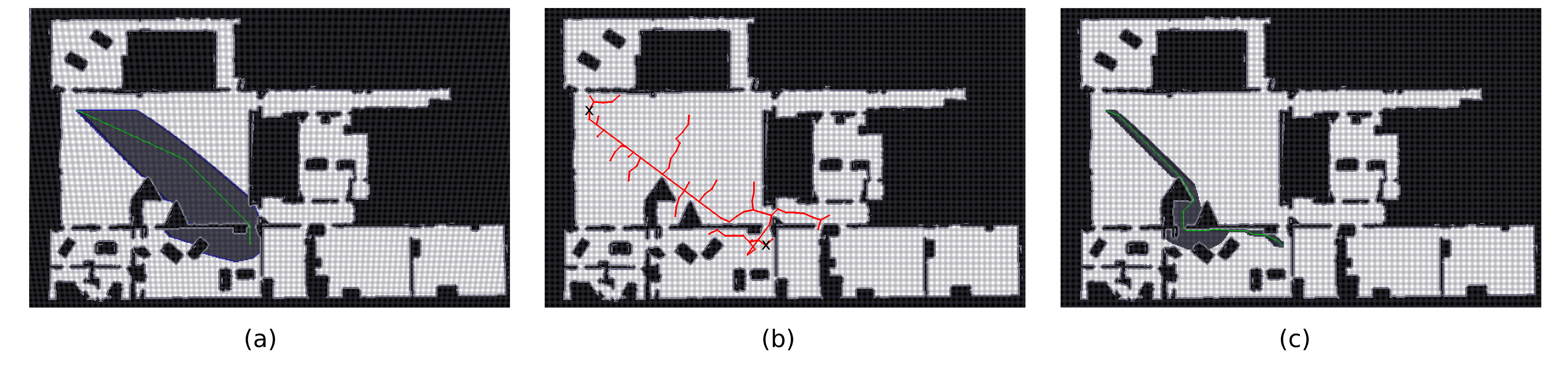}
    %\includegraphics[width=.47\columnwidth]{images/bar2.png}
    %\includegraphics[width=.47\columnwidth]{images/vio1.png}
    %\includegraphics[width=.45\columnwidth]{images/vio2.png}
  %\vspace{-4mm}
\caption{\SajadT{Visual comparison of planned paths on real world maps. a) A* b) d-RRT-Connect c) WPN (Note: Green represents the planned path. Grey is explored space and red is edges and nodes of sampling algorithms. Black Xs were used to denote the start and goal point for (b), since red edges cover the green path.)} }
  %\label{fig: wp_vs_a_1}
  \label{fig:realworld}  
  \vspace{-2 mm}
  \end{figure}

PathBench has the capability to natively interface with ROS and Gazebo to allow for seamless path planning for simulated and real world robotic applications. PathBench is able to visualize and plan for both fixed map environments and exploration environments. The planning is done in PathBench in real-time, and the control commands are sent to ROS to guide the robot. \SajadT{Fig.~\ref{fig:realworld} visualizes different algorithms' paths on real world maps acquired from SLAM performed in ROS.}
% The robot environment is simulated in Gazebo as a house style map, and the can also be visualized through Rviz, all launched from within PathBench. A sample experiment can be seen in Fig.~\ref{fig:gazebo_ros}. See the supplementary video and GitHub for more visualizations. 
% 
We can also demonstrate the live map capabilities of PathBench, using an algorithm with exploration capabilities. The robot can plan into known space, and also plan into the unknown environment, while the PathBench map is updated as it explores. This can also be seen in the supplementary video and GitHub, where the exploration is demonstrated.  

\vspace{-1 mm}
\section{Conclusion}
\label{sec:conclusion}
PathBench presents a significant advantage in terms of developing and evaluating classical and learned motion planning algorithms, by providing development environment and benchmarking tools with standard and customizable metrics. PathBench has been demonstrated across a wide range of algorithms, datasets \textcolor{black}{and systems with comparative studies using various hardware and planning environments in this paper. Further insights of algorithms, systems, and environments can be uncovered with extension to the PathBench framework.}

In the future, PathBench will be extended to allow benchmarking and training of additional learning-based algorithms, along with support for higher dimensional planning \textcolor{black}{with kinematic and dynamic constraints. We will also aim to integrate PathBench with notable simulated environments and planning datasets, such as iGibson and Habitat to offer more diverse and complex settings for benchmarking.}

\section*{Acknowledgements}
We acknowledge the technical contributions made by 
Judicael E Clair, 
Danqing Hu, 
Radina Milenkova, 
Zeena Patel, 
Abel Shields, and 
John Yao to the programming of the project and producing of
Fig.~\ref{fig: sim}-e-f, 
Fig.~\ref{fig: simulator3d}, and Fig.~\ref{fig:3maps}'s second row. Notable programming contributions made include the added support of PathBench's rendering and support of 3D path planning environments, infrastructural changes for efficiency and a renewed ROS interface.    

\section*{Funding}
This work was partially funded by DRDC-IDEaS (CPCA-
0126) and EPSRC (EP/P010040/1). 

%\section*{Disclosure Statement}
%No potential conflict of interest was reported by the authors.

\section*{Biographical Note}

\noindent {\bf Hao-Ya Hsueh} is a student at the Department of Mechanical and Industrial Engineering,  Ryerson University. He is a member of Robotics and Computer Vision Lab at Ryerson University. His research interests include robotics, path planning, and computer vision.\\

\noindent {\bf Alexandru-Iosif Toma} received M.Eng in Computing in 2019 from the Imperial College London. His research interests include machine learning, neural network, robotics, and computer vision. \\

\noindent {\bf Hussein Ali Jaafar} is a student at the Department of Mechanical and Industrial Engineering,  Ryerson University. He is a member of Robotics and Computer Vision Lab at Ryerson University. His research interests include mechatronics, computer vision, and robotics.\\

\noindent  {\bf Edward Stow} is a PhD Student in the Software Performance Optimisation Group at Imperial College London, having completed a M.Eng degree in Computing at Imperial College in 2020.\\

\noindent  {\bf Riku Murai} received M.Eng in Computing in 2019 from the Imperial College London. He is currently a PhD student in the Department of Computing at Imperial College London.
His research interests include robotics and computer vision. In particular, the use of novel hardware and distributed computations.\\

\noindent {\bf Paul H J Kelly} has been on the faculty at Imperial College London since 1989, has a BSc in Computer Science from UCL (1983) and has a PhD in Computer Science from the University of London (1987).  He leads Imperial’s Software Performance Optimisation research group, working on domain-specific compiler technology.\\

\noindent {\bf Sajad Saeedi} is an Assistant Professor at Ryerson University. He received his PhD in Electrical and Computer Engineering from the University of New Brunswick, Fredericton Canada. His research interests span over simultaneous localization and mapping (SLAM), focal-plane sensor-processor arrays (FPSP), and robotic systems.

\bibliographystyle{tfnlm}
\bibliography{sample}  % .bib

\clearpage

\end{document}